%% file: main.tex
\documentclass[10pt, journal, compsoc]{IEEEtran}
\usepackage{anyfontsize}
\usepackage{microtype}
\usepackage{etoolbox}
\usepackage[breaklinks=true, colorlinks, bookmarks=false]{hyperref}

\usepackage{xspace}
\usepackage{graphicx}
\usepackage{booktabs}
\usepackage{multirow}
\usepackage{color}
\usepackage{colortbl}
\usepackage{amssymb}
\usepackage{amsmath}
\usepackage{amsthm}
\usepackage{cite}
\usepackage{cleveref}
\usepackage[inline]{enumitem}

\usepackage[table]{xcolor}
\definecolor{mygray}{gray}{.92}
\definecolor{mycyan}{cmyk}{.3,0,0,0}
\definecolor{LightCyan}{rgb}{0.95,1,1}

\input{commands/newcommand}

\usepackage[norelsize, linesnumbered, ruled, lined, boxed, commentsnumbered]{algorithm2e}
\usepackage{xcolor}
\usepackage{pifont}

\definecolor{mygray}{gray}{.9}

\usepackage{array}
\newcolumntype{x}[1]{>{\centering\arraybackslash\hspace{0pt}}p{#1}}

\ifCLASSINFOpdf
\else
\fi

\begin{document}

\title{AI-Generated Images as Data Sources: The Dawn of the Synthetic Era}

\author{
    Zuhao~Yang,~
    Yu~Wei,~
    Fangneng~Zhan,~
    Kunhao~Liu,~
    Muyu~Xu,~
    Xiaoqin~Zhang,~
    Ling~Shao, \textit{Fellow}, \textit{IEEE},~
    Shijian~Lu$^{\S}$

\IEEEcompsocitemizethanks{
\IEEEcompsocthanksitem Z. Yang, Y. Wei, K. Liu, M. Xu and S. Lu are with the Nanyang Technological University, Singapore.
\IEEEcompsocthanksitem F. Zhan is with the Hong Kong University of Science and Technology, China Hong Kong.
\IEEEcompsocthanksitem X. Zhang is with Zhejiang University of Technology, China.
\IEEEcompsocthanksitem L. Shao is with the UCAS-Terminus AI Lab, UCAS, China.
\protect 
\IEEEcompsocthanksitem $\S$ denotes the corresponding author, E-mail: shijian.lu@ntu.edu.sg.
}
}

\markboth{IEEE TRANSACTIONS ON PATTERN ANALYSIS AND MACHINE INTELLIGENCE}
{Yang \MakeLowercase{\textit{\etal}}: AI-Generated Images as Data Sources: The Dawn of the Synthetic Era}

\IEEEtitleabstractindextext{
\begin{abstract}
The advancement of visual intelligence is intrinsically tethered to the availability of large-scale data. In parallel, generative Artificial Intelligence (AI) has unlocked the potential to create synthetic images that closely resemble real-world photographs.
This prompts a compelling inquiry: how much could visual intelligence benefit from the advance of generative AI?
This paper explores the innovative concept of harnessing these AI-generated images as new data sources, reshaping traditional modeling paradigms in visual intelligence.
In contrast to real data, AI-generated data exhibit remarkable advantages, including unmatched abundance and scalability, the rapid generation of vast datasets, and the effortless simulation of edge cases.
Built on the success of generative AI models, we examine the potential of their generated data in a range of applications, from training machine learning models to simulating scenarios for computational modeling, testing, and validation.
We probe the technological foundations that support this groundbreaking use of generative AI, engaging in an in-depth discussion on the ethical, legal, and practical considerations that accompany this transformative paradigm shift.
Through an exhaustive survey of current technologies and applications, this paper presents a comprehensive view of the synthetic era in visual intelligence. A project associated with this paper can be found at \href{https://github.com/mwxely/AIGS}{https://github.com/mwxely/AIGS}.
\end{abstract}

}

\maketitle

\IEEEdisplaynontitleabstractindextext

\IEEEpeerreviewmaketitle

\begin{figure*}[t]
    \centering
    \includegraphics[width=\linewidth]{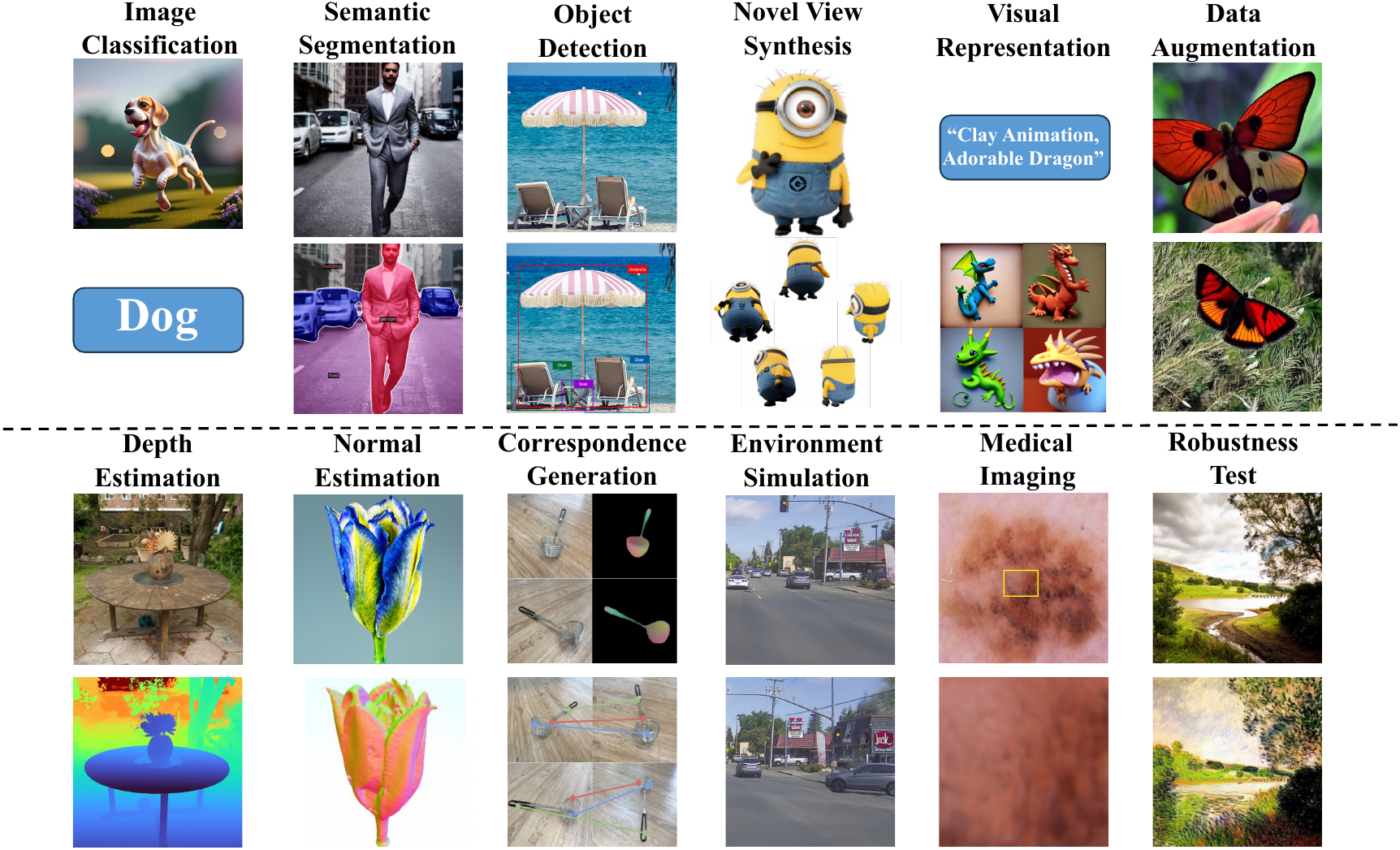}
    \caption{Illustration of AI-generated images as data sources. The AI-generated images could be utilized as training data in many widely explored computer vision tasks such as image classification, semantic segmentation, object detection, novel view synthesis, visual representation, data augmentation, etc. They could also be exploited in many more specific tasks such as depth estimation, normal estimation, correspondence generation, environment simulation for autonomous driving, medical imaging augmentation, robustness test, etc. The samples are from \cite{wu2023datasetdm, tian2023stablerep, burg2023data, liu2023zero, barron2022mip, wang2023prolificdreamer, yen2022nerfsupervision, yang2023unisim, bissoto2021gan, mofayezi2023benchmarking}. Best viewed when zoomed in.}
    \label{fig:teaser}
\end{figure*}

\IEEEraisesectionheading{\section{Introduction} \label{sec:intro}}
\IEEEPARstart{D}{ata} has been playing a crucial role in modern machine learning systems. Especially, systems that utilize deep learning models demand vast datasets to achieve good accuracy, robustness, and generalization.
However, the process of data collection, such as the manual annotation in various vision tasks, is often cumbersome and time-consuming.
Deep learning research is thus potentially hindered by a three-way dilemma, i.e., data quality, data scarcity, as well as data privacy and fairness \cite{lu2023machine}.
On the other hand, we have witnessed significant advancement of \textit{AI}-\textit{G}enerated \textit{C}ontent (AIGC) in producing highly photorealistic and diverse images. 
Such advancements in AIGC open up the fascinating possibility of replacing the real data with the inexhaustible AI-generated data, which enhances the controllability and scalability of data and mitigates the privacy concerns greatly \cite{joshi2022synthetic, yoon2018pategan, privacygan, Bozkir2019PrivacyPG}.
To this end, we investigate the concept of \textit{AI}-\textit{G}enerated images as data \textit{S}ources, termed AIGS, and provide profound insights about how the synthetic data produced by \emph{generative AI} could revolutionize the development of visual intelligence.

Synthetic data refers to the data generated by computer algorithms or simulations as an approximation of information gathered or measured in the real world \cite{app11052158, fernando2021realdeal}.
Prior to the explosion of AIGC, synthetic images were commonly generated with graphics engines or image composition.
For instance, the well-known Virtual KITTI \cite{gaidon2016virtual} is a dataset that was designed to learn and evaluate models for several video understanding tasks (e.g., object detection, multi-object tracking, instance segmentation). The authors leverage the off-the-shelf game engines to reconstruct scenes with automatically generated ground-truth labels.
Virtual KITTI 2 \cite{cabon2020virtual} is the updated version of Virtual KITTI with richer scene variants such as modified weather conditions and modified camera configurations, making it more suitable for benchmarking autonomous driving algorithms.
Composition-based synthetic images are widely adopted in computer vision tasks, especially in the areas of scene text detection and scene text recognition, providing extra samples for evaluating a model's generalization ability without manual annotation cost.
For example, \cite{gupta2016synthetic} and \cite{zhan2018verisimilar} propose to overlay the foreground text onto an existing background to form synthetic scene text images.
UnrealText \cite{long2020unrealtext} leverages a 3D graphics engine (i.e., Unreal Engine 4) to render text images and text in a 3D world. A two-stage pipeline is adopted to probe around object meshes and find proper text regions.
Both aforementioned approaches for synthetic image generation can greatly save annotation costs.
However, computer simulations with graphics engines fall short in generalizing to real-world datasets due to domain gaps \cite{chi2023datasetnerf}. In addition, image composition requires extra effort for visually understanding the correlation between foreground objects and background context.

AIGS methodologies, which bypass the tedious visual understanding process, directly produce high-quality and high-diversity images with smaller domain gaps.
In general, tools for visual content synthesis can be boiled down to two branches, namely, generative models and neural rendering.
Among generative models, Generative Adversarial Networks (GANs) \cite{goodfellow2014gan, mirza2014conditional, pix2pix2017, shrivastava2017learning, park2019semantic} and diffusion models (DMs) \cite{NEURIPS2020_4c5bcfec, song2020denoising, song2021scorebased, jolicoeur2020adversarial} have achieved notable success in image synthesis. Specifically, GANs can learn rich semantic information from latent space, enabling efficient image manipulation \cite{alaluf2022hyperstyle, ling2021editgan, shen2020interpreting}. DMs, as a new line of generative foundation models, have a stationary training objective and exhibit decent scalability \cite{zhan2021multimodal} to obtain better synthesis quality \cite{dhariwal2021diffusion}.
In addition to generative models, neural rendering \cite{mildenhall2020nerf} offers a valuable approach for synthesizing strictly multi-view consistent images from the learned 3D representation.

\begin{figure}[ht]
    \centering
    \includegraphics[width=\linewidth]{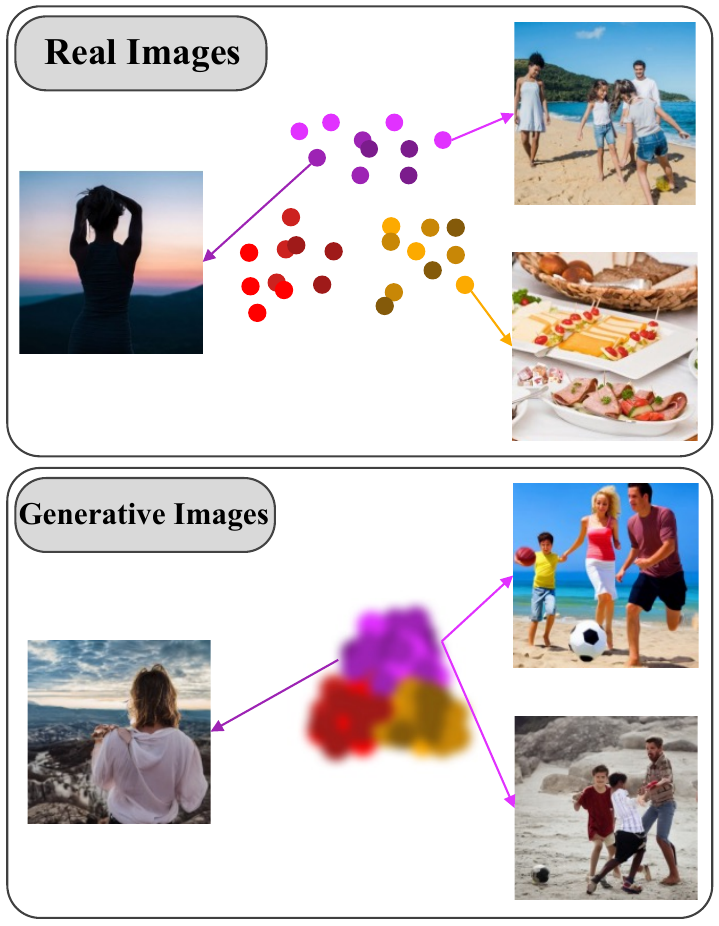}
    \caption{Illustration of generative images. Real images are sampled from discrete space, while generative models allow sampling images from a continuous data distribution. These images are sourced from \cite{tian2023stablerep}.}
    \label{fig:label_gen_data}
\end{figure}

{\righthyphenmin=2
AIGS methodologies leveraging generative models mainly encompass training data synthesis and testing data synthesis.
Synthetic training data originates from two sources, i.e., newly generated images with precise annotations, and supplementary images used for data augmentation.
While handling various downstream computer vision tasks, three approaches have been widely explored for acquiring the labels of synthesized images: 
\begin{enumerate*}[label=(\arabic*)]
    \item conditional generative models;
    \item latent space generalization;
    \item copy-paste synthesis.
\end{enumerate*}
With conditional generative models, annotations can be naturally obtained from conditional signals, especially for those training data used for image classification \cite{mirza2014conditional, odena2017acgan, brock2018large} and object detection \cite{chen2023integrating}.
As generative models enable the effective capture of semantic information from images through their rich latent code, segmentation masks of synthesized images can be generated with few manually annotated images \cite{zhang2021datasetgan, li2022bigdatasetgan, wu2023datasetdm} or refined cross-attention maps \cite{wu2023diffumask, xie2023mosaicfusion, nguyen2023dataset}. Besides, pyramid features from the U-Net can also be extracted to predict bounding boxes \cite{zhang2023diffusionengine}.
Recently, copy-paste synthesis has arisen as a novel pipeline for generating composite images with bounding box annotations. The foreground object is cut and pasted to the background image, and therefore the category and location of each foreground object become certain prior knowledge.
As for data augmentation, both fully-synthetic data as often produced by conditional generative models \cite{antoniou2017data, bowles2018gan, burg2023data} and semi-synthetic data as returned by semantic editing with latent space sampling (e.g., GAN inversion \cite{9792208, zhu2020indomain}) can be utilized to expand existing datasets.
Synthetic testing data has two primary use scenarios, i.e., generalization ability evaluation, and robustness test.
One can utilize synthetic images to form a more comprehensive testing set, leading to improved generalization evaluation of tested models \cite{makhmudkhujaev2021re, wu2023dynamic}.
In addition, as generative models are capable of translating images to another domain while preserving their semantics, domain-shifted synthetic images can be a promising data source for model robustness evaluations with less annotation cost \cite{mofayezi2023benchmarking}.
In this survey, we refer to those images produced by generative models as \emph{generative images}. The key difference between generative images and real images is illustrated in \Cref{fig:label_gen_data}.
\par}

With the emergence of neural rendering, particularly neural radiance fields (NeRF) \cite{mildenhall2020nerf} and 3D Gaussian Splatting (3DGS) \cite{kerbl20233d}, the computer vision community has shown growing interest in 3D-aware image synthesis. AIGS methodologies leveraging neural rendering mainly encompass 3D-aware training data synthesis and environment simulation. As shown in \Cref{fig:teaser}, numerous works have leveraged neural rendering to generate images with rich 3D annotations, including camera and object poses \cite{saxena2023generalizable, yen2021inerf, guo2022visual, avraham2022nerfels, Zhu2022LATITUDERG, Maggio2022LocNeRFMC, lewis2022narf22, lin2023parallel, matteo20246dgs,jin20256dope}, object correspondences \cite{yen2022nerfsupervision}, 3D bounding boxes \cite{li2023lift3d,cao20243dgs}, multi-view segmentation masks \cite{chi2023datasetnerf, liu2023weakly,cen2025segment}, meshes \cite{wang2021neus}, depths, and surface normals. The inherent multi-view consistency of neural rendering makes it particularly valuable for augmenting datasets in robotics \cite{yen2022nerfsupervision} and autonomous driving \cite{li2023lift3d, Tong20233DDA, li2023adv3d, zhou2024drivinggaussian}.

Both NeRF and 3DGS offer distinct advantages for synthetic data generation. NeRF's implicit neural representation excels at capturing complex scene geometry and appearance, making it well-suited for high-fidelity novel view synthesis in diverse environments. Meanwhile, 3DGS's explicit Gaussian representation enables direct association of semantic and geometric attributes with individual primitives, facilitating the rendering of richly annotated outputs such as segmentation masks, depth maps, and normals. Its faster training convergence and real-time rendering capabilities also make it attractive for large-scale dataset creation. In the context of autonomous driving, where corner case simulation is critical, both approaches provide promising solutions for environment and sensor simulation. By generating photorealistic and annotated data under diverse and safety-critical scenarios, neural rendering techniques are emerging as versatile foundations for constructing synthetic benchmarks and advancing 3D-aware perception systems.


To the best of our knowledge, this is the \emph{first} survey that comprehensively investigates the impact and enhancement of AI-generated images on various computer vision tasks and applications, together with extensive evaluation on generative images. Previously, \cite{app11052158} and \cite{math10152733} surveyed effective synthetic data generation, which respectively focused on generating synthetic images using non-deep learning techniques and GANs.
Joshi \emph{et al.} \cite{joshi2022synthetic} conducted a survey about the synthetic data for human-related applications and Man \emph{et al.} \cite{jimaging8110310} offered a general overview of the taxonomy of synthetic images and common methods for image synthesis, without emphatically discussing generative deep learning models like GANs or DMs and neural rendering tools.
In addition, several papers \cite{lu2023machine, wang2023security, wang2023survey} have investigated the security and privacy issues on generative data in AIGC.
Li \emph{et al.} \cite{li2023benchmarking} recently benchmarked the generative images for visual recognition tasks. Differently, we review the synthetic images as a data source by unifying the following three aspects:
\begin{enumerate*}[label=(\arabic*)]
    \item AIGS methodology formulation for generative models and neural rendering;
    \item AIGS application taxonomy for visual perception, visual generation, visual representation, as well as other domains involving computer vision (e.g., robotics and medicine);
    \item Evaluation of both the inherent quality of AI-generated images and how they benefit various downstream visual recognition tasks.
\end{enumerate*}

The contributions of this survey can be summarized as follows:
\begin{itemize}
    \item It encompasses extensive studies that explore AI-generated images as data sources, and embodies state-of-the-art AIGS technologies in a rationally structured framework.
    \item It presents fundamental ideas and background information of neural image synthesis and emphasizes how synthetic images are generated and utilized (\Cref{sec:nis}).
    \item It examines a broad array of AIGS applications in the realm of computer vision, such as visual perception tasks, visual generation tasks, and self-supervised learning (\Cref{sec:application}).
    \item It provides a summary of up-to-date synthetic datasets and evaluation metrics for generative images, benchmarking the improvements that generative images endow in terms of efficiency, cost, and performance with extensive qualitative and quantitative experiments (\Cref{sec:experiment}).
    \item It analyzes and discusses the social impact (\Cref{sec:social}) and challenges (\Cref{sec:challenge}) of AIGS, coupled with our humble insights on promising research directions and future development trends about AIGS.
\end{itemize}

\section{Methods} \label{sec:nis}
The core of AIGS methodologies lies with the utilization of generative models (\Cref{sec:gen_model}) and neural rendering (\Cref{sec:neu_red}).
In this section, we first discuss generative models, which allow learning data distributions from existing data for generating new data.
Thereafter, we introduce neural rendering, which serves as a promising approach to yield 3D-aware data.

\subsection{Generative Models} \label{sec:gen_model}
Broadly speaking, generative models include Generative Adversarial Networks (GANs) \cite{goodfellow2014gan, mirza2014conditional, pix2pix2017, shrivastava2017learning, park2019semantic}, variational autoencoders (VAEs) \cite{kingma2013auto}, autoregressive models \cite{menick2018generating, pmlr-v48-oord16}, flow models \cite{rezende2015variational, NEURIPS2018_d139db6a}, and diffusion models (DMs) \cite{NEURIPS2020_4c5bcfec}.
In particular, GANs and DMs stand out as the most common generative models used in AIGS due to their widespread adoption in visual generation.
In the following sections, we first revisit the fundamentals of GANs and DMs (\Cref{sec:gan}). After that, we present how GANs and DMs work for synthesizing useful training data (\Cref{sec:diff_model}).

\subsubsection{Generative Model Foundation} \label{sec:gan}

\textbf{Generative Adversarial Networks}. GANs have achieved tremendous success in synthesizing photorealistic images. They are typically composed of two neural networks: a generator network $G(z)$ aiming to generate synthetic images that have a distribution similar to that of real images $p_{\text{data}}$ and a discriminator network $D(x)$ that learns to distinguish between real images $x \sim p_{\text{data}}$ and the fake ones generated by $G(z)$ (the noise $z$ is sampled from a prior distribution $p_z$). These two networks are jointly optimized in a minimax manner, where the training objective can be formulated as follows:
\begin{equation}
    \begin{aligned}
        \min_G \max_D &\mathcal{L}(D, G) = \mathbb{E}_{x \sim p_{\text{data}}}[\log D(x)] \\
        &+\mathbb{E}_{z \sim p(z)}[\log (1-D(G(z)))],
    \end{aligned}
\end{equation}
where $\mathcal{L}(D, G)$ denotes the loss function that supervises this two-player minimax game.

Commonly used GANs include noise-to-image (N2I) translation GANs, image-to-image (I2I) translation GANs, and text-to-image (T2I) synthesis GANs. Notable N2I GANs include DCGAN \cite{RadfordMC15} that replaces the pooling layers with strided convolutions in the discriminator and fractional-strided convolutions in the generator, respectively; as well as WGAN \cite{Arjovsky2017WassersteinG} that employs the Earth Mover distance \cite{Rubner2000TheEM} minimization to offer more learning stability and speed up the training process while mitigating the mode collapse in vanilla GANs. The widely adopted I2I GANs encompass pix2pix \cite{pix2pix2017} that performs I2I translation tasks for paired training data and CycleGAN \cite{CycleGAN2017} for unpaired image translation. For T2I GANs, GAN-INT-CLS \cite{reed2016generative} presents the first attempt of incorporating text descriptions with an image generation pipeline using GANs. TAC-GAN \cite{dash2017tac} combines the aforementioned GAN-INT-CLS with AC-GAN \cite{odena2017acgan} to yield higher generation quality and diversity. GigaGAN \cite{kang2023gigagan} comes up with a novel large-scale (i.e., 1B-parameter) GAN architecture trained on the LAION2B-en \cite{schuhmann2022laionb} dataset for the T2I synthesis task.

\textbf{Diffusion Models.}
With the recent advances of denoised diffusion probabilistic models (DDPMs) \cite{NEURIPS2020_4c5bcfec, pmlr-v37-sohl-dickstein15}, DMs have gained popularity as a prevalent class of score matching \cite{swersky2011autoencoders, pmlr-v139-bao21b, song2021scorebased} generative models. Inspired by nonequilibrium thermodynamics \cite{pmlr-v37-sohl-dickstein15}, DDPM works with a forward diffusion process and a reverse diffusion process. The forward process can be seen as a Markov chain where the Gaussian noise $\epsilon_t \sim \mathcal{N}(0, \mathbf{I})$ is gradually injected with $t=1,\dots,T$ time steps to the original image $x_0 \sim p(x_0)$, where $\mathbf{I}$ denotes the identity matrix possessing the same dimensions as $x_0$, and $p(x_0)$ denotes the data distribution. The reverse process starts from time step $T$ and reverts the forward process iteratively to reconstruct images from the noise distribution. By denoting the transition process of the forward and reverse processes by $q(x_{t}|x_{t-1})$ and $p(x_{t-1}|x_{t})$, respectively, the forward process yields an isotropic Gaussian eventually:
\begin{equation}
    q(x_{t}|x_{t-1}) = \mathcal{N}(x_{t};\sqrt{1-\beta_{t}}x_{t-1}, \beta_{t}\mathbf{I}),
\end{equation}
where $\beta_{t}$ represents the variance schedule. The reverse process can be parameterized by:
\begin{equation}
    p_{\theta}(x_{t-1}|x_{t}) = \mathcal{N}(x_{t-1};\mu_{\theta}(x_t, t), {\Sigma_{\theta}(x_t, t)}^{2}\mathbf{I}),
\end{equation}
where $\theta$ denotes the learnable parameters during the reverse process, and $\mu_{\theta}(x_t)$ can be further expanded into a linear combination of the noisy image $x_t$ and the noise approximation model $\epsilon_{\theta}(x_t, t)$.

Due to the intractable nature of $p_\theta(x_0)$, we cannot directly compute its maximum likelihood objective. Instead, Ho \emph{et al.} \cite{NEURIPS2020_4c5bcfec} take inspiration from VAEs \cite{kingma2013auto} and reformulate the training objective through a variational lower bound of the negative log-likelihood of $p(x_0)$, as follows:
\begin{equation}
    \begin{aligned}
        \mathcal{L}_{VLB} = &\mathbb{E}_q[\underbrace{-\log p_\theta(x_0|x_1)}_{\mathcal{L}_0}+\underbrace{KL(q(x_T|x_0)\parallel p_\theta(x_T))}_{\mathcal{L}_T} \\
        &+\underbrace{\sum_{t=2}^{T}KL(q(x_{t-1}|x_{t},x_0)\parallel p_\theta(x_{t-1}|x_t))}_{\mathcal{L}_{t-1}}],
    \end{aligned}
    \label{eq:vlb}
\end{equation}
where $\mathbb{E}$ is the expected value, and $KL$ denotes the Kullback-Leibler divergence. It is worth noting that Ho \emph{et al.} \cite{NEURIPS2020_4c5bcfec} leverage a separate model to estimate $\mathcal{L}_0$ for better synthesis. $\mathcal{L}_T$ is a constant term, which eventually yields a simplified objective that can be formulated as follows:
\begin{equation}
    \begin{aligned}
        \mathcal{L}_t &= KL(q(x_{t-1}|x_{t},x_0)\parallel p_\theta(x_{t-1}|x_t)) \\
        &= \mathbb{E}_{t\sim[1,T],x_0\sim p(x_0),\epsilon_t\sim\mathcal{N}(0,\mathbf{I})}\parallel \epsilon_t-\epsilon_\theta(x_t,t)\parallel^2.
    \end{aligned}
\end{equation}
During the reverse process, a neural network is trained to predict the noise instead of estimating the mean and the covariance directly.

In addition, \cite{Croitoru2022DiffusionMI} claims that the DMs should be divided into three sub-categories:
\begin{enumerate*}[label=(\arabic*)]
    \item denoised diffusion probabilistic models (DDPMs) which we have just explained in detail; 
    \item noise conditioned score networks (NCSNs) \cite{NEURIPS2019_3001ef25} which leverage a shared neural network to approximate the score function (i.e., the gradient of the log density);
    \item stochastic differential equations (SDEs) \cite{song2021scorebased} which can be regarded as a generalization of the previous two modeling strategies but with stronger theoretical outcomes.
\end{enumerate*}

While DMs have a more stable training process and offer higher generation diversity \cite{dhariwal2021diffusion} thanks to their likelihood-based properties, GANs can be more efficient as they do not rely on multiple network evaluations during inference time \cite{zhan2021multimodal}.
In summary, generative images from both models can serve as promising data sources for various downstream tasks (e.g., classification \cite{Haque_2021, math10091541, 8363576, BIRD2022110684, jahanian2022generative, he2023is, tian2023stablerep, trabucco2023effective, burg2023data, azizi2023synthetic}, segmentation \cite{Tritrong2021RepurposeGANs, Chen_2019_CVPR, semanticGAN, 9956471, shaham2019singan, baranchuk2022labelefficient, wu2023diffumask, karazija2023diffusion}, detection \cite{9522995, 10.1007/s42979-023-01704-5, 9786867, lin2023fsod, ge2022dall, 10095167, voetman2023big, wu2022synthetic}) across multiple fields (e.g., medicine \cite{chen2021synthetic, tucker2020generating, 8363576, ali2023leveraging}, finance \cite{quantgan, tss_cgan, vuletic2023fin}, education \cite{studentgan, privacygan, etrgan}).

\subsubsection{Synthetic Data from Generative Models} \label{sec:diff_model}
The advancements of GANs and DMs empower the generation of photorealistic visual content \cite{Karras_2020_CVPR, Rombach_2022_CVPR}.
To incorporate these powerful generative models with AIGS, two core issues are:
\begin{enumerate*}[label=(\arabic*)]
    \item how to acquire the annotation of the generated images;
    \item how to employ the generated images to effectively augment existing datasets.
\end{enumerate*}
More details will be elaborated on in the ensuing subsections.

\textbf{Label Acquisition.}
Three major approaches have been adopted to acquire annotations of synthesized images, including conditional generative models, latent space generalization, and copy-paste synthesis.

\begin{figure}[ht]
    \centering
    \includegraphics[width=\linewidth]{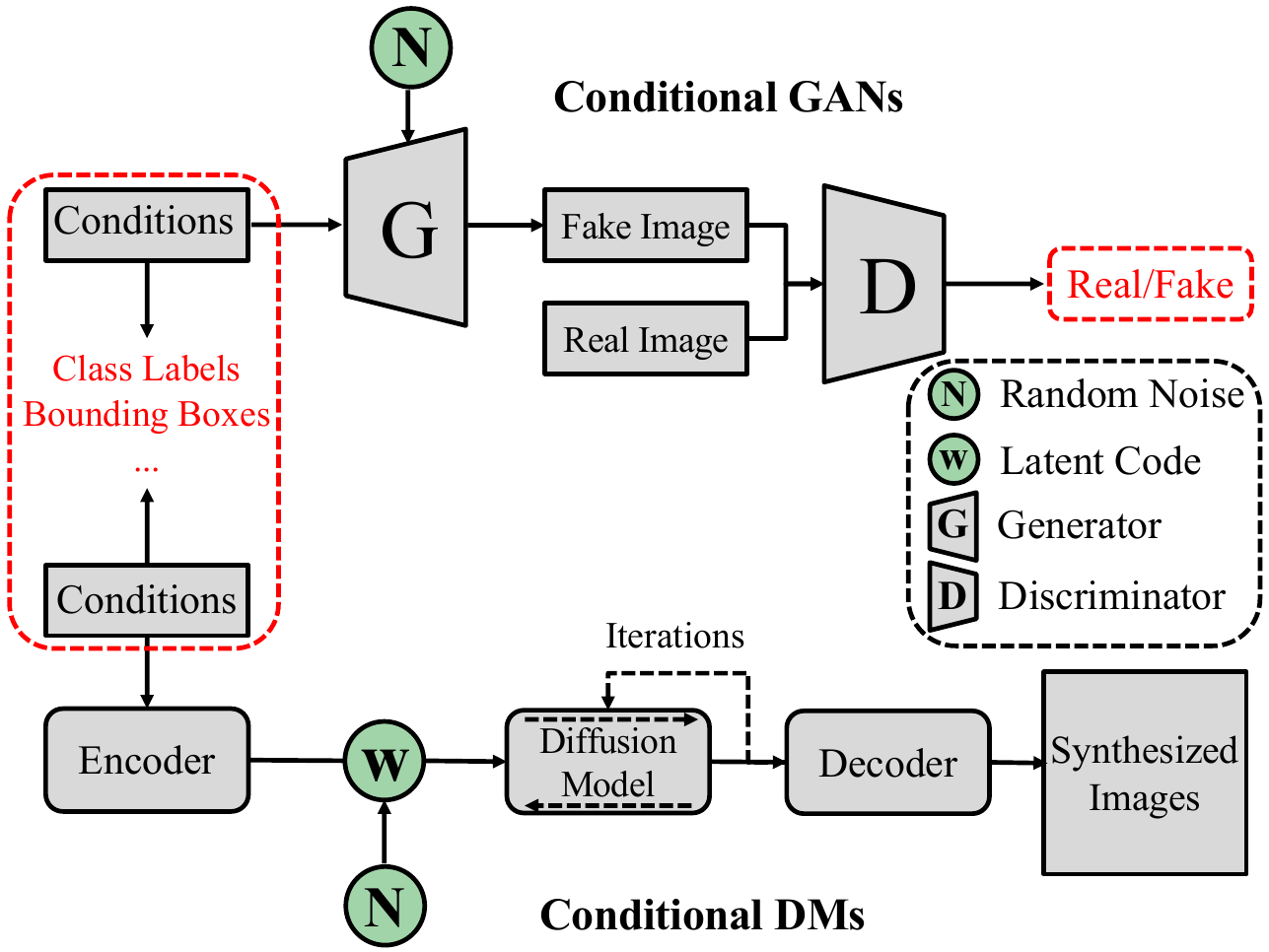}
    \caption{Illustration of label acquisition via conditional generative models. The basic architectures of conditional GANs and conditional DMs are provided. Solid rectangular boxes represent the network inputs. Rounded rectangular boxes denote the network components. The square box indicates the outputs.}
    \label{fig:label_cls}
\end{figure}

Conditional generative models \cite{mirza2014conditional, odena2017acgan, brock2018large, Rombach_2022_CVPR, saharia2022photorealistic, Ramesh2022HierarchicalTI, Nichol2021GLIDETP} provide significant convenience in obtaining class labels for downstream classification tasks. As illustrated in \Cref{fig:label_cls}, collecting class labels from class-conditioned generative models is intuitive, while retrieving class names from text-conditioned generative models requires querying specific templates based on predefined rules. In the context of object detection, layout-to-image generation \cite{chen2023integrating} that integrates geometric conditions to obtain bounding box annotations showcases impressive effectiveness under annotation-scarce circumstances.

\begin{figure}[ht]
    \centering
    \includegraphics[width=\linewidth]{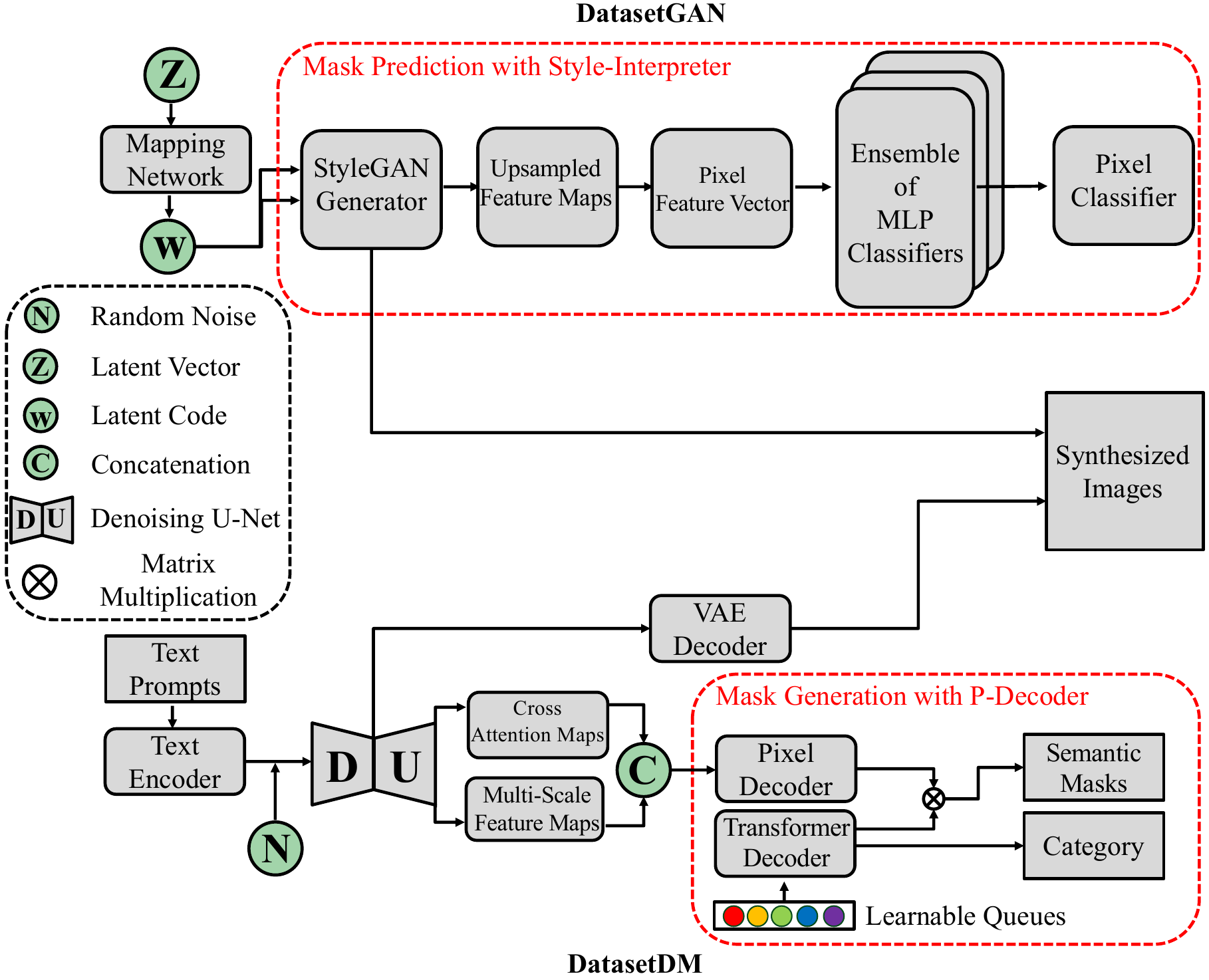}
    \caption{Illustration of label acquisition via latent space generalization. The basic architectures of DatasetGAN \cite{zhang2021datasetgan} and DatasetDM \cite{wu2023datasetdm} are provided. Solid rectangular boxes represent the network inputs. Rounded rectangular boxes denote the network components. The square box indicates the outputs.}
    \label{fig:label_seg}
\end{figure}

For latent space generalization, as generative models are capable of capturing rich semantic knowledge while synthesizing realistic images, several studies leverage the latent space of pre-trained GANs as feature interpreters for mask label prediction \cite{zhang2021datasetgan, li2022bigdatasetgan}. It is also valid to employ GAN inversion plus a shallow decoder for label generation \cite{Xu_2023_CVPR}. Both approaches have been acknowledged as groundbreaking contributions especially within the realm of image segmentation. Recently, this line of research is further pushed forward with the emergence of large-scale DMs. Specifically, to yield higher-quality annotations for synthetic images, \cite{wu2023datasetdm} utilizes the diffusion inversion, and achieves superior performance in multi-task scenarios (e.g., segmentation tasks, depth estimation, pose estimation). Several recent studies \cite{xie2023mosaicfusion, nguyen2023dataset} focus on acquiring segmentation masks via refining the cross-attention maps obtained from the DMs. \Cref{fig:label_seg} shows the architecture of mask acquisition in two dominant studies DatasetGAN \cite{zhang2021datasetgan} and DatasetDM \cite{wu2023datasetdm}.

\begin{figure*}[ht]
    \centering
    \includegraphics[width=0.9\linewidth]{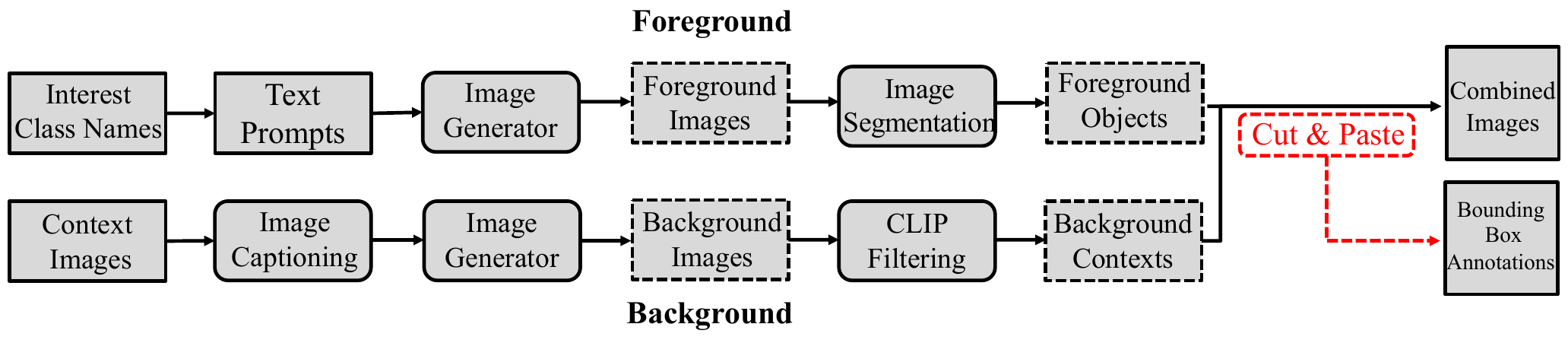}
    \caption{Illustration of label acquisition via copy-paste synthesis. The pipelines of foreground object generation and background context generation adopted in \cite{ge2022dall} are provided. Solid rectangular boxes represent the network inputs. Dotted rectangular boxes represent intermediate inputs. Rounded rectangular boxes denote the network components or image processing techniques. The square box indicates the outputs.}
    \label{fig:label_det}
\end{figure*}

For copy-paste-based synthesis,  \cite{lin2023fsod, ge2022dall, wu2022synthetic} separately generate foregrounds and backgrounds, and have become the prevalent detection data generation approach as illustrated in \Cref{fig:label_det}. The background generation can be easily achieved via T2I generative models, whereas the foreground objects are usually obtained by cropping \cite{lin2023fsod}, segmentation \cite{ge2022dall}, and a mask generator \cite{wu2022synthetic}.
As a special case in AIGS, the label-efficient generation paradigm \cite{baranchuk2022labelefficient, xu2023open, ni2023imaginarynet} has also attracted increasing attention recently.
For example, they exploit latent features of DMs for semantic segmentation \cite{baranchuk2022labelefficient, xu2023open}, and leverage synthetic imaginary images for object detection \cite{ni2023imaginarynet}.

\begin{figure*}[ht]
    \centering
    \includegraphics[width=\linewidth]{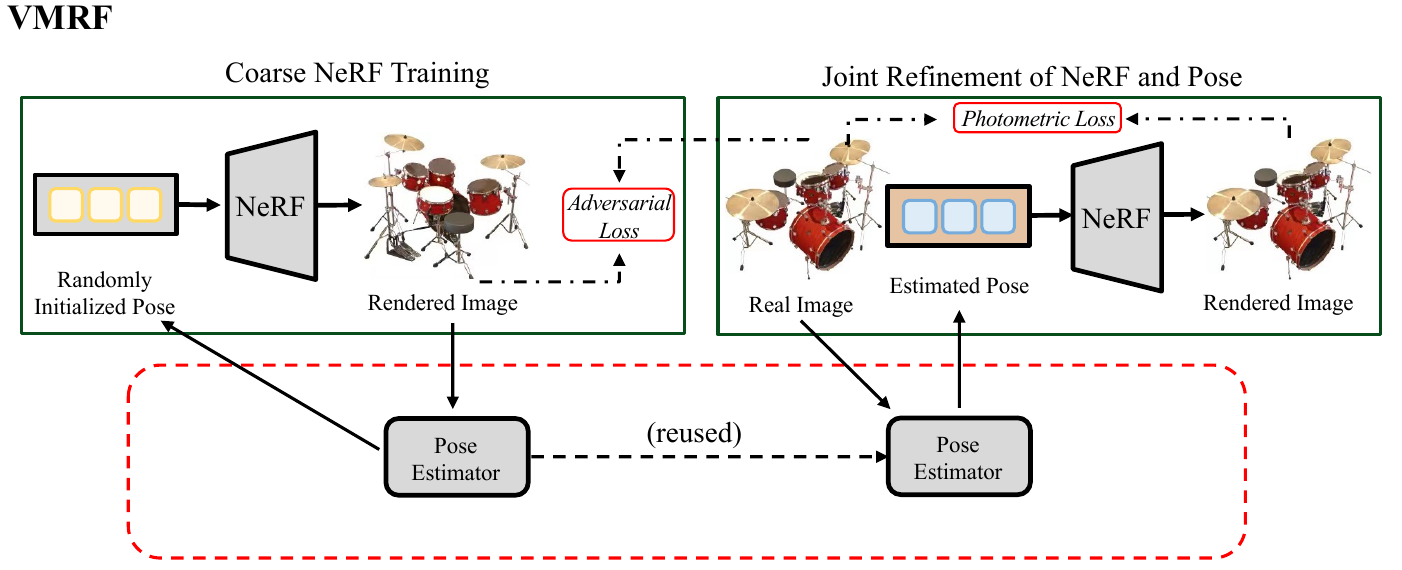}
    \caption{Illustration of label acquisition via neural rendering. The framework shows how VMRF \cite{zhang2022vmrf} estimates camera poses with rendered images and initialized poses. The trained pose estimator is reused to estimate poses of real images.}
    \label{fig:label_nerf}
\end{figure*}

Acquiring supervision labels via conditional generative models is intuitive as one can directly embed the annotation information using class names, text prompts, and geometric guidance, etc.
The other two label acquisition methods take inspirations from task-specific properties.
For instance, tasks like semantic segmentation and instance segmentation require rich semantic knowledge as priors, and hence one can utilize the latent features of GANs and DMs to efficiently synthesize segmentation masks.
Copy-paste synthesis is specially designed for detection-related tasks. The bounding box annotations can be easily tracked and retrieved since each foreground object is cut and pasted onto corresponding background context.

\textbf{Data Augmentation.} Generative images could also be leveraged to enlarge the size and enhance the diversity of existing datasets. Some popular GAN variants trained on large-scale datasets can be leveraged to generate high-resolution and superior-quality images for augmentation. For example, PGGAN \cite{karras2018progressive} produces CelebA \cite{liu2015faceattributes} images at 1024$\times$1024 pixels. BigGAN \cite{brock2018large} can be pretrained on ImageNet \cite{deng2009imagenet} images at 256$\times$256 and 512$\times$512. Style-based GANs \cite{Karras_2019_CVPR, Karras_2020_CVPR, NEURIPS2020_8d30aa96, NEURIPS2021_076ccd93} leverage adaptive instance normalization (AdaIN) \cite{Huang_2017_ICCV} to learn hierarchical latent styles and synthesize stylized FFHQ \cite{Karras_2019_CVPR} and LSUN \cite{yu15lsun} images with high perceptual quality.
In addition, \cite{burg2023data} reviews several categories of T2I DM-based augmentation methods, such as unconditional generation \cite{luzi2022blso} and 
generation with a single prompt or multiple prompts \cite{Radford2021LearningTV, sariyildiz2023fake}, joint DM optimization \cite{kawar2023imagic}, as well as pseudo-word representation \cite{gal2023an}.
Differently, \cite{trabucco2023effective} achieves augmentation with semi-synthetic (semantically edited) images where pre-trained T2I DMs \cite{Rombach_2022_CVPR} are exploited to alter high-level semantic attributes among the training data.
The aforementioned studies have empirically demonstrated that using DMs for data augmentation is effective and efficient as long as the real-fake ratio is carefully tuned.

\subsection{Neural Rendering} \label{sec:neu_red}

Generative models discussed in \Cref{sec:gen_model} focus on 2D image synthesis, without considering the 3D real-world information. In recent years, neural radiance field (NeRF) \cite{mildenhall2020nerf} has become a prevalent model for 3D-consistent novel view synthesis. More recently, 3D Gaussian Splatting (3DGS) \cite{kerbl20233d} has also emerged as a powerful approach, offering superior novel view synthesis performance with explicit 3D representations. We will go through the foundation of NeRF (\Cref{sec:nerf_found}) and 3DGS (\Cref{sec:3dgs_found}) followed by an elaborated explanation of how they can be used to generate synthetic images.

\subsubsection{NeRF Foundation} \label{sec:nerf_found}
NeRF achieves photorealistic rendering quality by modeling the color and the density of the 3D world. Specifically, given a 3D point  $\textbf{x}=(x,y,z)$ and 2D viewing direction $\textbf{d}=(\theta,\phi)$, a 5D vector-valued function $F_{\Theta}$ parameterized by $\Theta$ maps the point to an emitted color $\textbf{c}=(r,g,b)$ and volume density $\sigma$. The mapping function $F_{\Theta}$ can be an MLP \cite{mildenhall2020nerf, barron2021mip}, discrete voxel grids \cite{fridovich2022plenoxels, sun2022direct}, decomposed tensors \cite{fridovich2023k, chan2022efficient, chen2022tensorf}, hash maps \cite{muller2022instant}, etc. The color of each pixel can be computed by applying volume rendering \cite{drebin1988volume} along the ray $\textbf{r}$ emitted from the camera origin:
\begin{equation}
\begin{gathered}
\hat{\mathbf{C}}(\mathbf{r}) = \sum_{i=1}^{N} T(i)(1-\text{exp}(-\sigma(i)\delta(i)))\mathbf{c}(i),\\
T(i) = \text{exp}(-\sum_{i-1}^{j}\sigma(j)\delta(j)),
\end{gathered}
\end{equation}
where $\delta(i)$ represents the distance between two consecutive sample points along the ray, $N$ is the number of samples along each ray, and $T(i)$ indicates the accumulated transparency. 

Due to the differential property of the volume rendering process, NeRF can be used for diverse purposes. The most straightforward application of NeRF is 3D reconstruction for novel view synthesis. Given the multi-view images of a scene, the radiance field can be optimized by minimizing the reconstruction error between the rendered color $\hat{\mathbf{C}}$ and the ground truth color $\mathbf{C}$:
$
\mathcal{L} = \left \| \hat{\mathbf{C}}(\mathbf{r}) - \mathbf{C}(\mathbf{r}) \right \|_{2}^{2}.
$
With a reconstructed NeRF, we can apply multiple editing methods, including object maneuver \cite{yang2021learning, yuan2021star}, style transfer \cite{zhang2022arf, liu2023stylerf}, text-based editing \cite{haque2023instruct, yu2023edit}, and relighting \cite{martin2021nerf, rudnev2022nerf}, to get multitudinous outcomes.
NeRF can also be integrated into generative models \cite{chan2021pi, gu2021stylenerf, chan2022efficient, wang2023rodin, muller2023diffrf, anciukevivcius2023renderdiffusion}, enabling GANs and DMs to generate 3D assets. Such generative NeRFs are trained using generative objectives explained in \cref{sec:gen_model}. Additionally, NeRF can be used as a distillation objective for large-scale 2D foundation models to generate 3D objects from text descriptions \cite{poole2022dreamfusion, jain2022zero, wang2023prolificdreamer}.

\subsubsection{3DGS Foundation} \label{sec:3dgs_found}

3D Gaussian Splatting (3DGS) is a recent approach for neural rendering that represents 3D scenes as a collection of anisotropic 3D Gaussians. Unlike implicit representations such as NeRF, 3DGS employs an explicit representation where each Gaussian primitive is parameterized by its center position $\mu \in \mathbb{R}^3$, a 3D covariance matrix $\Sigma \in \mathbb{R}^{3 \times 3}$, an opacity value $\alpha \in [0,1]$, and view-dependent color encoded by spherical harmonic (SH) coefficients $c \in \mathbb{R}^{(l+1)^2 \times 3}$ (where $l$ is the maximum SH degree). The 3D Gaussian function is defined as:
\begin{equation}
G(x) = \exp\left(- \frac{1}{2}(x - \mu)^\top \Sigma^{-1}(x - \mu)\right),
\end{equation}
To ensure the positive semi-definiteness of $\Sigma$, it is parameterized as $\Sigma = RSS^\top R^\top$, where $S = \text{diag}(s)$ with $s \in \mathbb{R}^3_+$ represents the scaling factors, and $R \in SO(3)$ is the rotation matrix derived from a unit quaternion $q \in \mathbb{R}^4$.

During rendering, 3D Gaussians are projected onto the 2D image plane through an affine approximation of the projective transformation. Given a viewing transformation $W$ and a Jacobian $J$ of the affine approximation of the projective transformation, the 2D covariance $\Sigma'$ in camera space is computed as:
\begin{equation}
\Sigma' = JW\Sigma W^\top J^\top.
\end{equation}
The projected 2D Gaussians are then rasterized using a tile-based approach, where Gaussians overlapping each tile are sorted by depth and blended via alpha compositing. For a pixel, the final color $C$ is computed by blending $N$ ordered Gaussians from front to back:
\begin{equation}
C = \sum_{i=1}^{N} c_i \alpha'_i \prod_{j=1}^{i-1} (1 - \alpha'_j),
\end{equation}
where $c_i$ is the color evaluated from SH coefficients based on the viewing direction, and $\alpha'_i = \alpha_i G_i(x)$ combines the learned opacity $\alpha_i$ with the 2D Gaussian weight $G_i(x)$ evaluated at pixel location $x$. This differentiable splatting-based rendering enables end-to-end optimization of all Gaussian parameters through gradient descent, while achieving real-time performance through efficient rasterization. The explicit nature of the representation also facilitates direct manipulation of scene properties, making 3DGS particularly suitable for tasks requiring semantic or geometric annotations.

\subsubsection{Synthetic Data from Neural Rendering} \label{sec:syn_nerf}
As neural rendering naturally induces multi-view consistency and cross-view correspondence, the synthetic images obtained from NeRF and 3DGS can be used for many 3D-aware applications such as pose estimation, tracking, detection, correspondence, navigation, etc.
3D-aware training data synthesis with neural rendering involves label acquisition and data augmentation as well, which we will elaborate 
in the ensuing two subsections.

\textbf{Label Acquisition.} 
Iterative generation is the most commonly adopted procedure for AIGS with neural rendering. By incorporating a temporal state $s$, the corresponding NeRF or 3DGS $G$ can generate synthetic images conditioned on $s$: $\hat{I} = G(s)$. The synthetic image $\hat{I}$ will be used to compute a target loss function. The gradients are back-propagated to the temporal state $s$ rather than the networks $G$. The temporal state $s$ is updated in every iteration whereas the generator $G$ is frozen. The output of the algorithm will be $s$ in the last iteration, and the trained NeRF or 3DGS is only used as a conditional generative model to evaluate the current state.
Iterative generation can be adopted for a range of applications, including camera and object pose estimation \cite{saxena2023generalizable, yen2021inerf, guo2022visual, avraham2022nerfels, Zhu2022LATITUDERG, Maggio2022LocNeRFMC, lewis2022narf22, lin2023parallel,fu2024colmap, wei2025pcr}, robot navigation \cite{adamkiewicz2022vision}, and tracking \cite{masuda2023event}. While most of them use neural rendering to generate RGB images, several methods try to generate features \cite{Moreau2023CROSSFIRECR, Chen2023RefinementFA}, events \cite{masuda2023event}, and occupancy \cite{Chen2023CATNIPSCA}.

Common labels in 3D data include: camera pose, object correspondence, multi-view segmentation masks, mesh, normal, and depth, each of which requires a different approach to obtain.
Camera poses can be obtained through joint optimization with neural rendering. For example, CF-3DGS~\cite{fu2024colmap} jointly refines camera poses and Gaussian parameters by enforcing correspondence-aware consistency between Gaussian-rendered images and real observations.
To obtain accurate correspondence labels from neural rendering, one can perform correspondence generation via a distribution of depths, which works well when the density distribution is multimodal along the ray \cite{yen2022nerfsupervision}.

Furthermore, \cite{chi2023datasetnerf} harness an enhanced semantic tri-plane \cite{chan2022efficient} and a semantic feature decoder to aggregate and interpret 3D features before volumetric rendering, yielding highly consistent multi-view segmentation masks.
A mesh can be easily retrieved from a NeRF representation via the marching cubes algorithm \cite{lorensen1987marching}, which can then be used to obtain normal maps.
Depth maps can be naturally acquired since both NeRF and 3DGS can render consistent RGB-D images~\cite{zhang2025cdgs}. These neural rendering methods can effectively function as depth sensors, providing a single-valued depth at each discrete pixel from multiple views. Please refer to \Cref{fig:label_nerf} for the illustration of acquiring camera poses.

\textbf{Data Augmentation.} Both NeRF and 3DGS are capable of generating images from any novel view, making them valuable tools for augmenting multi-view datasets. There are several applications using neural rendering for data augmentation, such as semantic segmentation \cite{liu2023weakly}, object detection \cite{Ge2022NeuralSimLT}, navigation \cite{Tong20233DDA, Chen2023CATNIPSCA}, camera pose estimation \cite{Moreau2021LENSLE, meng2021gan}, object pose estimation \cite{Li2022NeRFPoseAF}, and object correspondence \cite{yen2022nerfsupervision}.
NeRF and 3DGS serve as abundant sources of 3D-consistent multi-view images, especially with advancements in large-scale 3D generative models \cite{Skorokhodov20233DGO, Xiang20233DawareIG,long2024wonder3d}. The use of the generated images from neural rendering methods as a data source for downstream tasks is a promising and ongoing area of exploration.

\section{Applications} \label{sec:application}

\begin{figure*}[t]
    \centering
    \includegraphics[width=0.9\linewidth]{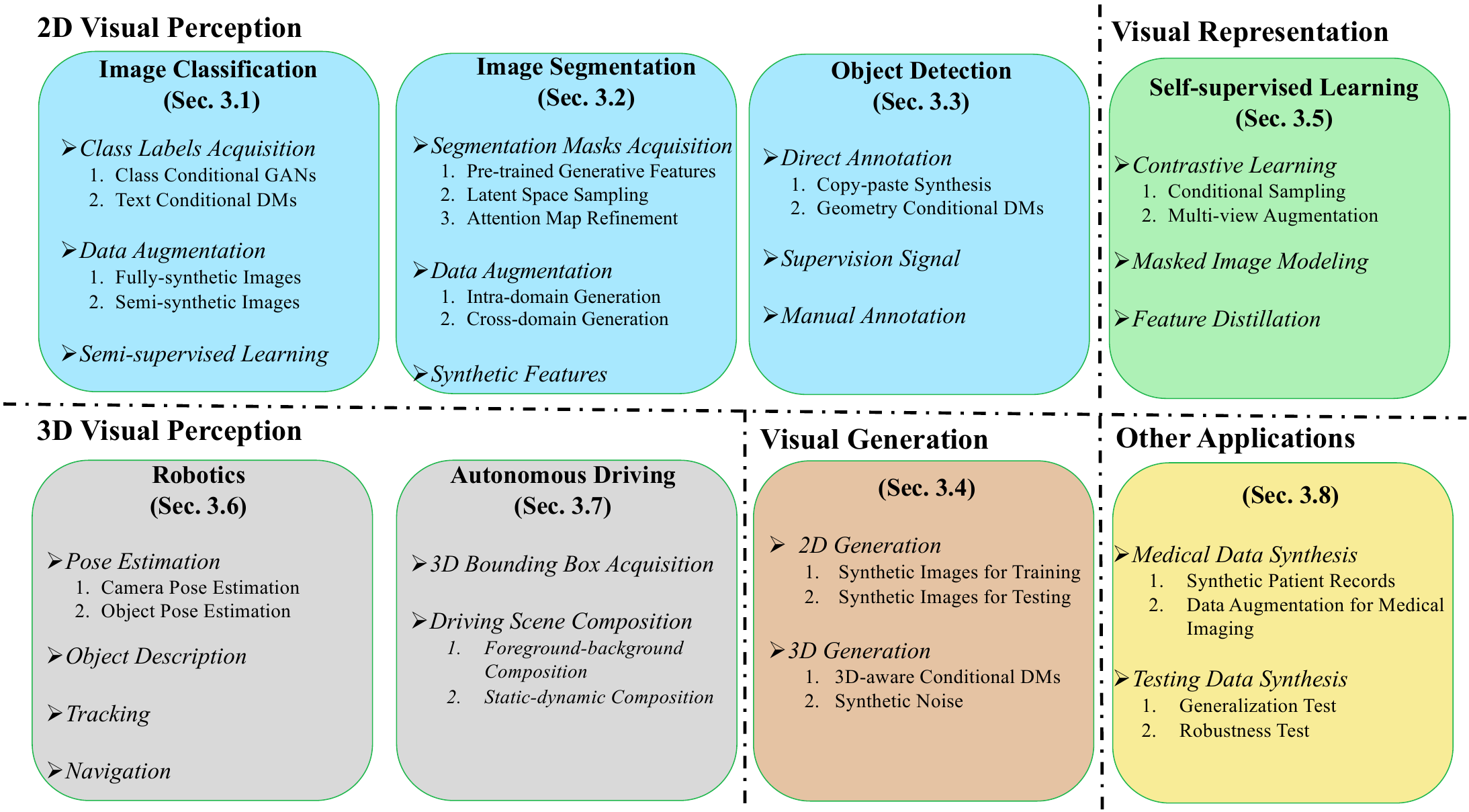}
    \caption{Taxonomy of AIGS applications. The synthetic images could be utilized as training data or augmentation data in 2D visual perception tasks such as image classification, semantic segmentation, object detection, etc. They could also be exploited in visual representation, visual generation, and other specific applications such as medical data synthesis and testing data synthesis. Synthetic images with 3D-aware annotations are particularly useful for tasks in the realms of robotics and autonomous driving such as pose estimation, drone navigation, driving scene composition, etc.}
    \label{fig:app}
\end{figure*}

AIGS has empowered various downstream computer vision tasks. The AIGS-related applications can be broadly grouped into five categories:
\begin{enumerate*}[label=(\arabic*)]
    \item 2D visual perception tasks, including image classification (\Cref{sec:cls}), image segmentation (\Cref{sec:seg}), and object detection (\Cref{sec:det}) that command the majority share of AIGS applications so far;
    \item visual generation tasks (\Cref{sec:gen}), where synthetic images are employed in training generative models rather than discriminative ones;
    \item self-supervised learning tasks (\Cref{sec:self}), where synthetic images are exploited to train visual representation learners;
    \item 3D visual perception tasks, including applications across robotics (\Cref{sec:rob}) and autonomous driving (\Cref{sec:auto}) domains, where synthetic images can be employed to transmit 3D information when modeling the complex 3D scenes;
    \item other applications (\Cref{sec:others}), where synthetic images are adopted in some specific domains (e.g., medical and model testing).
\end{enumerate*}
A detailed taxonomy is presented in \Cref{fig:app}.


\subsection{Image Classification} \label{sec:cls}
Image classification is a fundamental task in the field of computer vision, with the aim of teaching machines to comprehend and categorize visual data just as human vision does. Leveraging advanced machine learning algorithms and deep neural networks, image classification enables computers to recognize patterns, objects, and features within images, assigning them to predefined classes or categories. By extracting meaningful features from the input images and learning from vast labeled datasets, the system becomes capable of identifying and distinguishing various objects and concepts with impressive accuracy.

Synthetic images serve two main purposes in image classification: replacing the original training set or augmenting the existing training set.
As an early attempt, Besnier \emph{et al.} \cite{Besnier2019ThisDD} employ BigGAN \cite{brock2018large} to generate more diverse and informative data via latent code optimization, showcasing encouraging results on ImageNet classification with solely GAN-generated training images but on a small scale (i.e., 10 classes).
OpenGAN \cite{9710170} generalizes to unbounded open-set data via training an open-vs-closed classifier on off-the-shelf features computed by the closed-world K-way classification network rather than pixels.
Recently, several studies \cite{he2023is, sariyildiz2023fake, zhou2023training} utilize T2I DMs to generate text-aligned training images and mitigate the issues of low diversity and non-stationary training objectives rooted in GAN-based methods. In particular, He \emph{et al.} \cite{he2023is} perform extensive experiments to show that synthetic images produced by GLIDE \cite{Nichol2021GLIDETP} can remarkably boost the performance of zero-shot and few-shot learning. Sariyildiz \emph{et al.} \cite{sariyildiz2023fake} refine the text prompts with WordNet \cite{wordnet} information and auxiliary backgrounds to reduce the semantic ambiguity. Furthermore, Zhou \emph{et al.} \cite{zhou2023training} implement diffusion inversion to obtain the latent vectors corresponding to real images and then sample novel images based on the learned noises.
All aforementioned studies use synthetic images as a novel data source for training classifiers and empirically demonstrate clear performance improvements of the trained models.

More studies focus on augmenting the existing datasets instead of using the synthetic images solely, especially in the realms of medicine and healthcare as personal data is scarce and private.
For example, GAN-based data augmentation has been widely adopted for medical image classification \cite{8857905, FRIDADAR2018321, 8363576}. Recently, Stable Diffusion \cite{Rombach_2022_CVPR} is harnessed to generate synthetic radiological images according to domain-specific prompts \cite{chambon2022adapting}.
In general, there are four approaches to augment classification datasets.
Firstly, \cite{yuan2022not, bansal2023leaving, shipard2023diversity} utilize diverse self-curated prompt templates to guide the image generation and accomplish promising results on zero-shot shifted-distribution classification.
Secondly, Lei \emph{et al.} \cite{lei2023image} and Dunlap \emph{et al.} \cite{dunlap2023diversify} employ the captioning model to construct prompt sets for image synthesis, which both use instance-level prompts instead of class-level information to further enhance the diversity of generated images.
Thirdly, several studies fine-tune the image generator to achieve better downstream performance. For example, Azizi \emph{et al.} \cite{azizi2023synthetic} fine-tune Imagen \cite{saharia2022photorealistic} on ImageNet-1K, while Kawar \emph{et al.} \cite{kawar2023imagic} jointly optimize the conditioning vector and the diffusion U-Net.
Last but not least, Gal \emph{et al.} \cite{gal2023an} and Shin \emph{et al.} \cite{shin2023fill} employ Textual Inversion \cite{gal2023an} that samples prompt templates and jointly optimizes a pseudo-word that represents the class-concept.

The abovementioned methods mainly exploit fully synthetic image generation in a fully-supervised manner. As a complement to these approaches, \cite{trabucco2023effective, jahanian2022generative} use semantically edited (semi-synthetic) images produced from GAN inversion\cite{9792208} and an I2I DM to augment existing datasets. Furthermore, You \emph{et al.} \cite{you2023diffusion} validate that augmentation using generative images is viable for semi-supervised classification through a three-stage dual pseudo training paradigm.

To conclude, image classification is utilized in a multitude of domains, encompassing autonomous vehicles, medical imaging, facial recognition, and content-based image retrieval. This technology is revolutionizing our interactions with visual information and fostering numerous innovative applications in contemporary society.
Leveraging synthetic images in image classification tasks exhibits favorable advantages including but not limited to high-fidelity image production, reduced storage employment, as well as unbounded generation quantity.

\subsection{Image Segmentation} \label{sec:seg}
In image processing and computer vision, image segmentation involves dividing an image into distinct regions or objects, comprising sets of pixels with similar semantic labels. The primary objective of image segmentation is to simplify or transform the image representation into a more meaningful and analytically tractable form. Image segmentation is commonly employed to identify objects and boundaries, such as lines and curves within the image. More precisely, it involves assigning a label to each pixel in an image so that pixels with the same label exhibit certain shared characteristics. Synthetic image synthesis for image segmentation can be boiled down to two approaches. The first approach involves generating synthetic images using latent codes or other modalities, such as text, depth, or lines, through generative models. The second approach involves augmenting existing labeled data.

Despite the existence of numerous large-scale public datasets, such as PASCAL~\cite{everingham2010pascal}, COCO~\cite{lin2014microsoft}, and Cityscapes~\cite{cordts2016cityscapes}, the collection of per-pixel annotations for such datasets is costly and labor-intensive. With advancements in generative AI, synthetic images serve as a more cost-effective data source to train semantic segmentation models. For example, several studies~\cite{li2021semantic, Tritrong2021RepurposeGANs} utilize GAN-based models, such as StyleGAN2~\cite{Karras_2020_CVPR}, to generate synthetic images from latent codes, followed by manual intervention or automatic model-based annotation to obtain segmentation masks.
Additionally, \cite{mondal2018few, souly2017semi, jain2021using} utilize synthetic images as fake patches for adversarial training. With the rise of T2I generative models, some research explores using textual descriptions instead of solely relying on latent codes to generate a large volume of synthetic images for training models and recognizing different segmentation regions referenced in the text. For instance, \cite{karazija2023diffusion} utilizes Stable Diffusion to generate numerous images of cats and dogs, training segmentation models to segment cats and dogs within the same image. Li \emph{et al.} \cite{li2023openvocabulary} also employ a DM to create synthetic images, and leverage a readily available object detector~\cite{liu2021swin} to produce accurate oracle ground-truth masks. \cite{kondapaneni2023text} harnesses the CLIP model \cite{Radford2021LearningTV} besides the DM model to align the image caption and domain caption. Building upon this foundation, Gong \emph{et al.} \cite{gong2023prompting} further propose that utilizing the DM as the backbone network for segmentation offers a feature representation that is resilient to significant domain shifts, thereby enhancing generalization across diverse domains. Consequently, they leverage a backbone network comprising a pre-trained diffusion model augmented with a segmentation head to generate visually similar images across distinct domains from textual prompts, along with their corresponding segmentation maps.

Some studies focus on augmenting limited training sets using generative models instead of creating entirely new synthetic datasets. In the domains where large-scale datasets are scarce, such as the medical field \cite{ali2023leveraging}, data augmentation via generative models becomes imperative.
The data augmentation is largely achieved via two major approaches.
The first approach generates a substantial amount of data with similar or identical labels to the existing segmentation labels within the same domain. For instance, \cite{choi2019self} employs LSGAN-based \cite{mao2017least} data augmentation to generate enhanced data of the same label space in the target domain.
\cite{li2023intra} maps the original images to the latent space of StyleGAN2 \cite{Karras_2020_CVPR}, which can reproduce augmented images that effectively improve the performance of numerous segmentation models.
\cite{fawakherji2021multi, oh2023diffmix, tang2023multi} utilize label-based generation of corresponding fake synthetic images for adversarial training. While FreeMask \cite{yang2024freemask} directly utilizes a DM to generate diverse real images from the same semantic maps, some work like \cite{tan2023diffss, jia2023dginstyle, schnell2023generative} employs ControlNet \cite{zhang2023adding} to generate more precise auxiliary images based on the same segmentation map or the scribble label maps, for training a few-shot segmentation model.
MosaicFusion \cite{xie2023mosaicfusion} presents a training-free diffusion-based augmentation pipeline that simultaneously generates synthetic images via a T2I DM and corresponding masks via aggregation over cross-attention maps.
The second approach first manipulates the existing segmentation maps and then synthesizes the new maps.
For example, \cite{liu2019pixel} decomposes various regions in the existing segmentation map into sub-regions of different classes, then assembles these sub-regions to create new complete segmentation maps. It uses Pix2pix HD \cite{wang2018pix2pixHD} to generate new images based on these assembled segmentation maps.
\cite{fernandez2022can} employs a pre-trained latent diffusion model (LDM) \cite{Rombach_2022_CVPR} to generate similar segmentation maps based on existing maps and then adopts SPADE \cite{park2019semantic} to create new synthetic images for subsequent model training. SegGen \cite{ye2023seggen} proposes a pipeline consisting of an image captioner, a text-to-mask model, and a mask-to-image generation model to achieve the augmentation of their data. For each real image, it can generate diverse masks based on the caption of that image and synthesize the corresponding new images.

As a special case in AIGS for segmentation, recent work \cite{xu2023open} aims to leverage intermediate features from DMs to predict segmentation masks, without directly using the generative labeled images. It manages to extract diffusion features from a frozen T2I diffusion U-Net whose features can then be transmitted into a mask generator for class-agnostic binary mask predictions.

In conclusion, image segmentation is essential for digital image processing and computer vision, enabling the identification of objects and boundaries within images. To overcome challenges posed by costly and labor-intensive data collection, synthetic image generation with generative models offers a valuable alternative to either replace the original training source or enrich the existing datasets. 
These approaches open up new avenues for advancing semantic segmentation and promise more efficient and accurate visual data analysis across various applications, benefiting computer vision research as a whole. Continuous exploration and refinement of these methods hold great potential for the future of image segmentation as well as other related areas.

\subsection{Object Detection} \label{sec:det}
Object detection is a crucial field within computer vision that focuses on automated identification and localization of objects of interest within digital images or video streams. By leveraging sophisticated algorithms and deep learning techniques, object detection systems can accurately identify and draw bounding boxes around various objects, even in complex and cluttered scenes.
AIGS for object detection involves two key aspects: 
\begin{enumerate*}[label=(\arabic*)]
    \item detection data generation where obtaining supervision labels is a central concern;
    \item data augmentation with generative images. We will elaborate on the two aspects in the ensuing text.
\end{enumerate*}

The generation of detection data has been explored in three typical approaches, depending on how the supervision information is obtained.
The first approach achieves direct bounding box annotations via the AIGS methodologies proposed in \ref{sec:intro}.
For example, Lin \emph{et al.} \cite{lin2023fsod} apply saliency detection on generated images, and then crop novel instances to form composite images with accurate bounding box annotations. Likewise, Ge \emph{et al.} \cite{ge2022dall} separate their pipeline with foreground generation and context background generation, both of which leverage DALL-E \cite{pmlr-v139-ramesh21a} and Stable Diffusion, followed by the background-foreground segmentation to obtain foreground object masks.
Both abovementioned studies present their efficient copy-paste synthesis pipeline for generating a labeled detection dataset.
Differently, Chen \emph{et al.} propose GeoDiffusion \cite{chen2023integrating}, a layout-to-image conditional DM, to encode geometric conditions (e.g., bounding boxes, camera views) via text prompts and fine-tune the DM to generate auto-labeled images.
In addition, DiffusionEngine \cite{zhang2023diffusionengine} proposes a single-stage framework for detection data scaling up, which effectively harnesses LDM features for detection adaptor optimization.
The second approach involves indirect supervision signals where no bounding box labeling is required during training. For instance, Ni \emph{et al.} \cite{ni2023imaginarynet} propose an Imaginary-Supervised paradigm for training a detection model, where a representation generator is incorporated to extract proposal representations from imaginary images generated by a T2I synthesis model. The detection head with proposal representations and pre-sampled class labels can then be optimized with neither real images nor human annotations.
The third approach utilizes generative models for detection image generation but with manual labeling. For example, Voetman \emph{et al.} \cite{voetman2023big} use DreamBooth \cite{ruiz2023dreambooth} to fine-tune the output of a Stable Diffusion network, replicating images of a certain scenario and employing manual bounding box annotation to curate for their training set.

Several GAN-based studies focus on the data augmentation for object detection. For example, Martinson \emph{et al.} \cite{martinson2021training} propose a novel three-layer framework that employs CycleGAN \cite{CycleGAN2017} for synthetic imagery translation to complement the utility of 3D models, making a breakthrough of the annotation barrier, especially for rarely occurring objects. Kim \emph{et al.} \cite{9786867} train various detection networks and prove that, in the infrared small target detection scenario, using both real images and GAN-generated images exhibits better performance than using real images alone.
Besides, neural rendering-based approaches also show effectiveness in the data augmentation for object detection. For instance, Ge \emph{et al.} present the Neural-Sim \cite{Ge2022NeuralSimLT} pipeline that finds optimal parameters for generating views that can be used as the synthetic training data for downstream detection, empirically demonstrating the performance of their method on the ``YCB-in-the-Wild'' \cite{Ge2022NeuralSimLT} benchmark.

To summarize, object detection finds widespread applications across numerous domains, revolutionizing how machines interact with and interpret the visual world around them. As object detection continues to evolve, it holds the potential to reshape industries, improve safety, and drive innovation across countless sectors. 
With newly raised AIGS approaches for detection data generation and augmentation, it requires less human effort when training the detection model while possessing the comparably decent performance.

\subsection{Visual Generation} \label{sec:gen}


\begin{figure}[ht]
    \centering
    \includegraphics[width=0.75\linewidth]{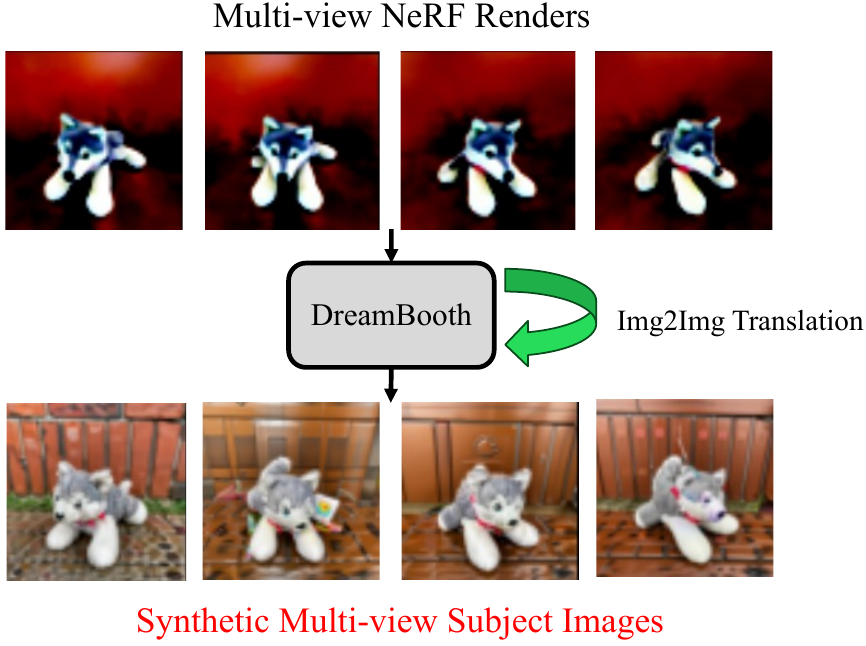}
    \caption{Example framework of AIGS for 3D generation tasks. The figure shows the multi-view data generation pipeline of DreamBooth3D \cite{raj2023_dreambooth3d}, where Dreambooth \cite{ruiz2023dreambooth} is adopted to synthesize subject-driven pseudo images at the I2I translation stage.}
    \label{fig:label_gen_2}
\end{figure}

Going beyond the realm of recognition and detection, visual generation harnesses the power of advanced generative models, such as GANs and DMs, to produce realistic images, videos, and even 3D assets. By learning from extensive datasets and patterns inherent in visual media, these generative models can generate original and compelling content, spanning diverse domains like art, design, animation, and even medical imaging. In this section, we discuss examples that focus on incorporating generative images during the training of generative models.

In the domain of 2D generation, several studies aim to use GAN-synthesized datasets for face age transformation during training and testing.
For example, Zoss \emph{et al.} \cite{zoss2022production} propose a production-ready face re-aging network (FRAN) that samples training images from the preceding Style-based Age Manipulation method \cite{alaluf2021only}.
Furthermore, Makhmudkhujaev \emph{et al.} present Re-aging GAN \cite{makhmudkhujaev2021re} that leverages the generated face images of StyleGAN2 \cite{Karras_2020_CVPR} to test the generalization capability of their model.

In the realm of 3D generation, several attempts leverage DMs to produce intermediate multi-view images for 3D generation and reconstruction thanks to the all-encompassing priors of appearance and geometry offered by the state-of-the-art conditional DMs.

For instance, Liu \emph{et al.} propose Zero-1-to-3 \cite{liu2023zero}, a framework that generates images from a specified camera viewpoint given only a single RGB image. Tang \emph{et al.} propose LGM~\cite{tang2024lgm}, a 3D Gaussian Splatting framework in which 3D models can be generated from single-view images or text prompts. In particular, LGM leverages diffusion models (DMs) to synthesize multi-view images, and an asymmetric U-Net is used to predict and fuse Gaussians for high-resolution 3D generation. Two hash grids are employed to reconstruct geometry and appearance from Gaussian renderings, with Marching Cubes extracting a coarse mesh that is iteratively refined via differentiable rendering. To bridge the domain gap between rendered training views and DM-synthesized inference views, LGM introduces Grid Distortion and Orbital Camera Jitter, improving robustness to view inconsistencies and camera pose inaccuracies.

Text-guided DMs have demonstrated great potential in text-to-3D generation. For instance, Tang \emph{et al.} propose GaussianDreamer~\cite{tang2023dreamgaussian} that utilizes diffusion-based supervision, where a pre-trained text-to-image diffusion model guides the optimization of Gaussian attributes through score distillation sampling (SDS). Specifically, the 3D Gaussians are initialized from a coarse point cloud generated by 3D diffusion models, upon which operations such as noisy point growing and color perturbation are applied to enrich the initialization. The Gaussian representation is then iteratively optimized under SDS guidance, benefiting from the geometry priors of both the 3D diffusion model and the 3D Gaussian Splatting framework.


In addition, Zhang \emph{et al.} propose StyleAvatar3D \cite{Zhang2023StyleAvatar3DLI}, where they employ ControlNet \cite{zhang2023adding} to produce multi-view images with pose guidance (e.g., depth maps, human pose images) obtained from a software engine as well as the style guidance obtained from predefined text prompts. All the three aforementioned studies exploiting generative images present superior performance over previous methods in terms of the quality and the diversity of visual generation.

To summarize, visual generation technology possesses immense potential, fueling creativity, enhancing realism in virtual environments, and pushing the boundaries of what is visually possible. Prepare to be awe-struck by the wonders of visual generation, where artificial intelligence harmoniously blends imagination and reality in astonishing harmony.
Utilizing synthetic images for visual generation tasks greatly enhances the efficiency and flexibility during the generation process, meanwhile unlocking more diversified outputs.

\subsection{Self-supervised Learning} \label{sec:self}


\begin{figure}[ht]
    \centering
    \includegraphics[width=0.75\linewidth]{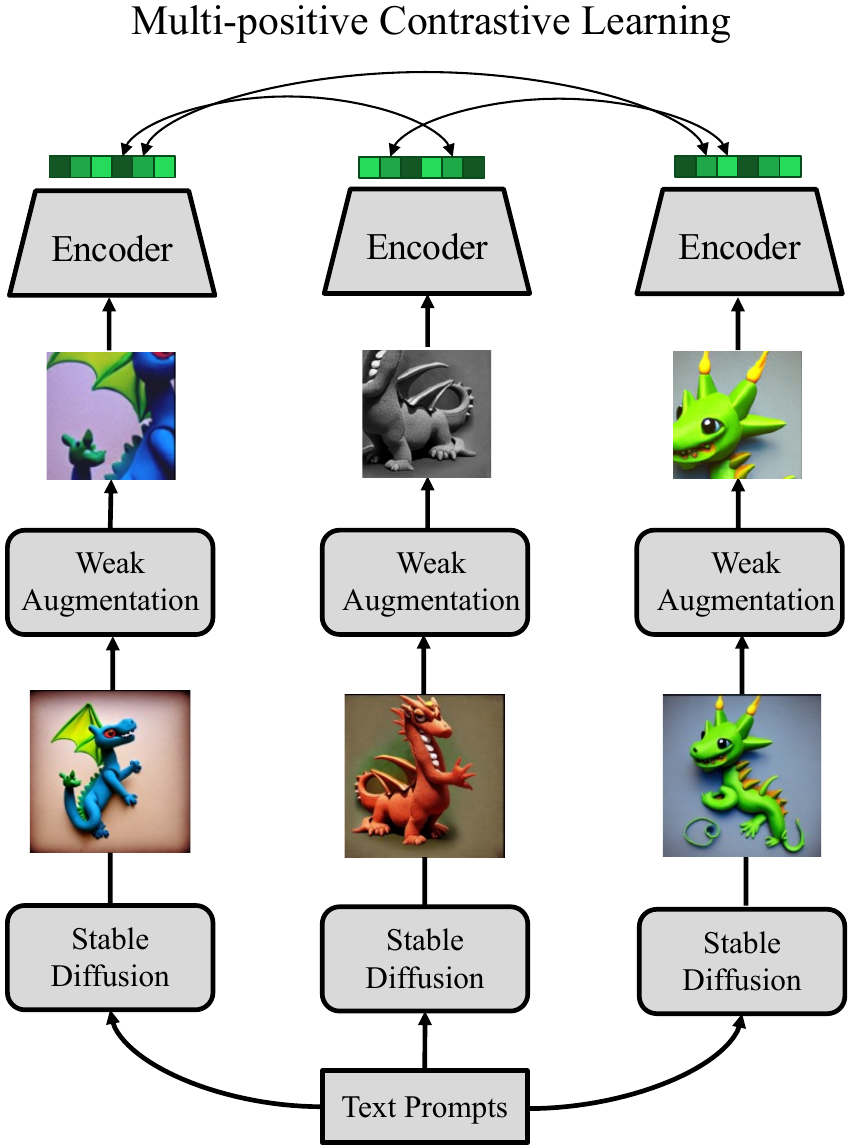}
    \caption{Pipeline of StableRep \cite{tian2023stablerep}. Synthetic images generated from Stable Diffusion are augmented and treated as positives for each other. The learning process is supervised by a multi-positive contrastive loss.}
    \label{fig:label_sr}
\end{figure}

Self-supervised learning is an innovative and promising approach that addresses the challenge of acquiring large-scale labeled datasets required for traditional supervised learning methods. Unlike conventional techniques that heavily rely on human annotation, self-supervised learning leverages the inherent structure and information within the data itself to generate supervision signals. By formulating tasks that require the model to predict certain properties or transformations of the input data, the system learns to extract meaningful representations and features autonomously. This revolutionary paradigm has demonstrated remarkable success in various downstream tasks, such as image recognition, object detection, and segmentation. 

Self-supervised learning algorithms can be broadly categorized into two families: 
\begin{enumerate*}[label=(\arabic*)]
    \item contrastive learning that contrasts the positive pairs with negative pairs of the same image in embedding space, with SimCLR \cite{chen2020simple} being the representative;
    \item masked image modeling that leverages unmasked patches to predict masked patches, with MAE \cite{he2022masked} being the representative.
\end{enumerate*}
Jahanian \emph{et al.} \cite{jahanian2022generative} and Chai \emph{et al.} \cite{chai2021ensembling} enhance self-supervised representation learning by using generative multi-view images. They both utilize latent manipulation in GANs to create augmented image pairs for contrastive learning, demonstrating that learning visual representations from implicit generative models can significantly improve performance compared to learning from pixel-space translation alone.
Tian \emph{et al.} extended their experimental context to both SimCLR and MAE, demonstrating that their newly proposed StableRep \cite{tian2023stablerep} pipeline (\Cref{fig:label_sr}), which utilizes Stable Diffusion to generate synthetic images for data augmentation, leads to a clear performance boost compared to training on real images alone.
SynCLR \cite{tian2024learning} further leverages both synthetic captions and synthetic images for pretraining, competing favorably with other general-purpose visual representation learners such as CLIP \cite{Radford2021LearningTV} and DINO v2 \cite{oquab2023dinov2}.
As a special case of AIGS for self-supervised learning, Li \emph{et al.} \cite{li2023dreamteacher} distill learned features and obtained labels from a pre-trained generative model into a target image backbone to perform downstream ImageNet classification, exhibiting superior performance without leveraging the synthetic images directly.

To conclude, with self-supervised learning's ability to capitalize on vast amounts of unlabeled data, self-supervised learning holds great potential to unlock new possibilities in visual understanding, significantly reducing data annotation efforts and advancing the frontiers of artificial intelligence in vision-based applications.
Harnessing synthetic images to create large-scale and diverse datasets for training has been empirically shown to be a promising enhancement for self-supervised representation learning, all without the need for manual annotation.

\subsection{Robotics} \label{sec:rob}

\begin{figure*}[ht]
    \centering
    \includegraphics[width=0.75\linewidth]{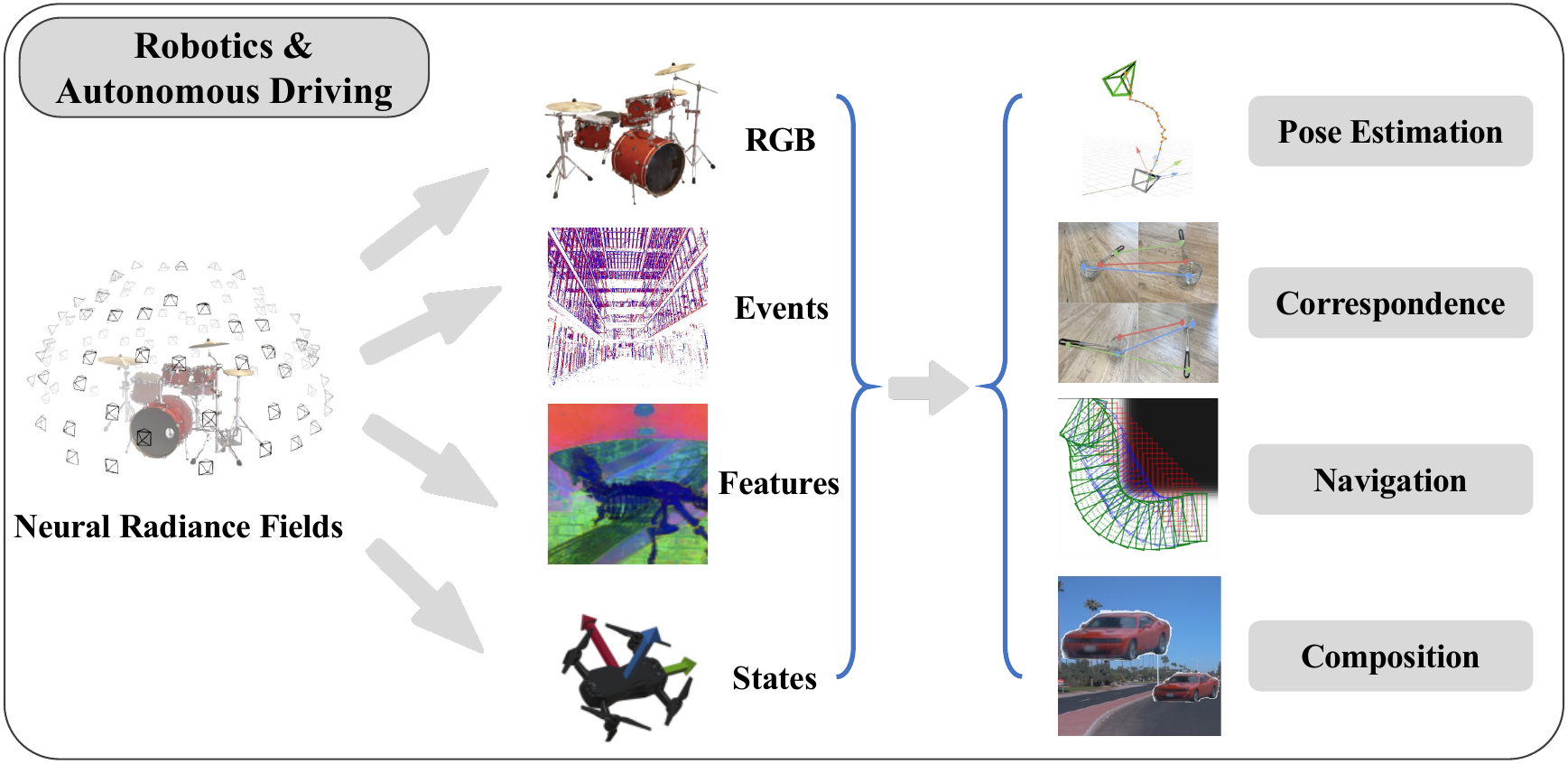}
    \caption{Basic framework of AIGS for robotics and autonomous driving. Multi-modality outputs such as RGB images, events, features, and states can be synthesized while NeRF is being optimized. These intermediate products can be harnessed to serve for various downstream tasks including but not limited to camera pose estimation, object correspondence, drone navigation and driving scene composition.}
    \label{fig:label_rob}
\end{figure*}

Robotics in computer vision is a captivating field that empowers machines with the ability to interpret and comprehend visual data, much like the human visual system.
With the recent advancement of neural rendering, AIGS greatly promotes the research in robot navigation \cite{adamkiewicz2022vision} and tracking \cite{masuda2023event}. When it comes to the data augmentation perspective, rendered images are also effective for camera pose estimation \cite{yen2021inerf, zhang2022vmrf, Moreau2021LENSLE}, object pose estimation \cite{Li2022NeRFPoseAF}, as well as object correspondence \cite{yen2022nerfsupervision}.

For example, 
Lin \emph{et al.} propose NeRF-Supervision \cite{yen2022nerfsupervision}, a novel RGB-sensor-only pipeline for learning object descriptors from synthetic dense correspondence, overcoming the recognition difficulties in the NeRF-based robotics context as NeRF does not directly estimate the boundaries of objects.

Parallel to NeRF-based works, recent 3DGS-based works offer compelling advantages for robotics applications, in particular for simulation and data augmentation. For instance, Splat‑Nav~\cite{chen2025splat} presents a real-time robot navigation pipeline built on Gaussian-splat maps, enabling fast trajectory planning and vision-based localization at update rates markedly higher than NeRF-based methods.
Meanwhile, EmbodiedSplat~\cite{chhablani2025embodiedsplat} demonstrates how 3D Gaussian splat reconstructions of deployment environments can be used to generate large-scale simulation environments for embodied agent training and strong sim-to-real transfer. 

These efforts highlight how 3DGS supports simulation of realistic environments and data amplification for robotics training pipelines — by providing fast, editable, explicit scene representations that are well-suited for augmentation, novel-view synthesis, and virtual policy fine-tuning.

\subsection{Autonomous Driving} \label{sec:auto}

Autonomous driving represents a transformative frontier in computer vision technology. It encompasses the development of advanced algorithms and deep learning models that enable vehicles to perceive and interpret their surroundings, making real-time decisions to navigate safely.

Autonomous driving datasets are usually collected with multi-view cameras that cover the full 360° field of view around the vehicle. To synthesize datasets with high-resolution images and accurate 3D labels, Li et al. propose Lift3D \cite{li2023lift3d} that provides accurate 3D bounding box annotations for downstream tasks by lifting a well-disentangled 2D GAN to a 3D object NeRF. Their method effectively boosts the performance of 3D object detectors on augmented autonomous driving datasets.

As a precise sensor simulator plays a crucial role in solving corner cases, several recent studies \cite{yang2023unisim, wu2023mars} focus on the environmental sensor simulation. Specifically, Wu et al. \cite{wu2023mars} propose to utilize two NeRFs to model the foreground instances and background environments separately. All the scene and object samples are composed together before volume rendering.

Recent work, however, shows that 3DGS is becoming a compelling alternative for driving-scene reconstruction and simulation. For example, DrivingGaussian~\cite{zhou2024drivinggaussian} presents a composite Gaussian splatting framework which incrementally builds a static background with 3D Gaussians and separately reconstructs dynamic objects (vehicles, pedestrians) using a Gaussian-graph representation, and benefits from LiDAR priors to enhance geometry and multi-camera consistency. 
Furthermore, SplatAD~\cite{hess2025splatad} leverages 3DGS for photorealistic driving-scene simulation and large-scale data augmentation, effectively enriching training data for autonomous driving systems. It constructs realistic virtual environments with controllable conditions such as viewpoints, lighting, and weather, enabling flexible generation of diverse scenarios for perception and planning tasks. Street Gaussians \cite{yan2024street} extend 3D Gaussian Splatting to large-scale driving scenes by decomposing static backgrounds and dynamic objects into Gaussian primitives. This explicit representation enables efficient and photorealistic rendering with temporal consistency, making it well-suited for autonomous driving simulation. Its high efficiency and object-level controllability further facilitate realistic scene composition and sensor simulation for data augmentation and testing.

These studies suggest that 3DGS offers efficient, high-fidelity explicit scene representation and real-time rendering—hence it is well-positioned to serve the needs of autonomous driving simulation and data augmentation. By integrating 3DGS into the dataset synthesis and simulation pipeline for autonomous driving, one can envision the following advantages: more accurate scene geometry, higher rendering efficiency, explicit object-level control (for dynamic vehicles/pedestrians), and seamless support for multi-modal sensors (camera + LiDAR). 
\Cref{fig:label_rob} shows an overall framework regarding AIGS for robotics and autonomous driving.

\subsection{Other Applications} \label{sec:others}
In this section, we will briefly discuss some specific yet equally important AIGS applications. In particular, we will concentrate on the potential positive influence of synthetic data within medical and model testing areas.

In the light of privacy concerns in the medical field, collecting relevant information can be challenging and even impossible, and synthetic data has thus drawn tremendous attention as a feasible solution.
For example, as a pioneering work, MedGAN \cite{pmlr-v68-choi17a} is designed to produce high-dimensional discrete variables based on real patient records.
Torfi \emph{et al.} propose CorGAN \cite{Torfi2020CorGANCC} to generate healthcare records by capturing correlations between adjacent medical features in the representation space.
Several studies focus on the augmentation perspective. For instance, GAN-based image augmentations are adopted for skin lesion \cite{8857905} and liver lesion \cite{8363576, FRIDADAR2018321} classification.
Ali \emph{et al.} \cite{ali2023leveraging} analyze 43 studies that reported GAN models for COVID-19 synthetic data generation and pinpoint the shortcomings and future directions of GAN-based augmentation for the medical imaging research community.

Synthetic images also play a valuable role in evaluating computer vision models, especially for benchmarking a model's generalization ability and robustness.
For instance, in the domain of face re-aging \cite{makhmudkhujaev2021re} where testing data across a continuous sequence of face ages is needed, an expressive image generator \cite{Karras_2020_CVPR} can produce photorealistic unseen images thereby creating a more comprehensive testing set for evaluating the model's generalization capability.
In the field of 3D scene reconstruction, Wu \emph{et al.} utilize synthetic datasets, as proposed in D-NeRF \cite{pumarola2021d}, to assess the performance of their model when modeling monocular dynamic scenes.
Moreover, in the realm of deepfake detection \cite{wang2023benchmarking}, synthetic images can also be harnessed as an abundant data source for evaluation since state-of-the-art generative models excel in producing diversified negative (fake) samples.
Regarding robustness tests, synthetic images generated from text-conditioned generative models have been proven to be effective and efficient in tuning more robust image classifiers, since synthesized domain-shifted images are capable of presenting more informative discrepancies with less effort compared to using hand-picked data for benchmarking \cite{mofayezi2023benchmarking}.



\section{Experimental Evaluation} \label{sec:experiment}

\subsection{Synthetic Datasets}
Datasets are the essence of computer vision tasks. Thanks to the rise of AIGS, existing scarce datasets can be augmented with samples of a decent degree of diversity in both content and style while saving the annotation cost at the same time.
In practice, synthetic datasets can be categorized into two subgroups:
\begin{enumerate*}[label=(\arabic*)]
    \item fully-synthetic datasets which are completely composed of generative images. They are usually curated for multi-modal visual understanding tasks (e.g., DiffusionDB \cite{wang-etal-2023-diffusiondb}) where each generative image is paired with a corresponding text prompt, human preference prediction tasks \cite{wu2023human}, regional editing detection tasks \cite{sun2023rethinking}, as well as subject-driven generation tasks \cite{zhou2024toffee}.
    \item semi-synthetic datasets (e.g., ForgeryNet \cite{he2021forgerynet}, DeepArt \cite{wang2023benchmarking}, GenImage \cite{zhu2023genimage}) which contain both real images and generative (fake) images. They typically have similar numbers of real images and fake images. Such datasets are well-suited for evaluating detection-based applications such as deepfake face detection, deepfake artwork detection, general-purpose image detection, etc.
\end{enumerate*}
Please refer to \Cref{tab:summ_datasets} for more detailed specification. Some synthetic datasets like GTAV \cite{Richter_2016_ECCV} and NeRF-Synthetic \cite{mildenhall2020nerf} are commonly adopted as data sources for training computer vision models. However, their images or views are produced from a 3D graphics engine rather than neural image synthesis.

\begin{table*}[t]
    \centering
    \caption{Summarization of popular datasets with synthetic images. The upper part denotes the detailed information of fully-synthetic datasets. The lower part denotes the information regarding semi-synthetic datasets. Note that for each semi-synthetic dataset, the real-fake ratio rather than the size of the whole dataset is tabulated. Part of the information is sourced from \cite{zhu2023genimage}.}
    \resizebox{0.9\linewidth}{!}{
        \begin{tabular}{l|c|c|c|c|c}
            \toprule[2pt] \textbf{Dataset} & \textbf{Application} & \textbf{Generator Type} & \textbf{Public Availability} & \textbf{Size (\emph{real} : \emph{fake})} & \textbf{Year} \\
            \midrule\midrule
            DiffusionDB \cite{wang-etal-2023-diffusiondb} & Prompt Inversion & DM & $\checkmark$ & 14 million & 2022 \\
            JourneyDB \cite{pan2023journeydb} & \begin{tabular}{@{}@{}c@{}@{}}Prompt Inversion, \\ Style Retrieval, \\ Image Captioning, \\ Visual Question Answering\end{tabular} & DM & $\checkmark$ & 4,692,751 & 2023 \\
            Pick-a-Pic \cite{kirstain2023pick} & Human Preference Prediction & DM & $\checkmark$ & 583,747 & 2023 \\
            ImageReward \cite{xu2023imagereward} & Human Preference Prediction & DM & $\checkmark$ & 354,608 & 2023 \\
            HPD v2 \cite{wu2023human} & Human Preference Prediction & DM & $\checkmark$ & 867,520 & 2023 \\
            GRE \cite{sun2023rethinking} & Regional Editing Detection & GAN \& DM & $\times$ & 228,650 & 2023 \\
            Toffee \cite{zhou2024toffee} & Subject-driven Generation & DM & $\checkmark$ & 5,400,000 & 2024 \\
            \midrule
            UADFV \cite{8683164} & Deepfake Face Detection & GAN & $\times$ & 241 : 252 & 2019 \\
            FakeSpotter \cite{Wang2019FakeSpotterAS} & Deepfake Face Detection & GAN & $\times$ & 6,000 : 5,000 & 2019 \\
            DFFD \cite{on-the-detection-of-digital-face-manipulation} & Deepfake Face Detection & GAN & $\checkmark$ & 58,703 : 240,336 & 2020 \\
            APFDD \cite{gandhi2020adversarial} & Deepfake Face Detection & GAN & $\times$ & 5,000 : 5,000 & 2020 \\
            ForgeryNet \cite{he2021forgerynet} & Deepfake Face Detection & GAN & $\checkmark$ & 1,438,201 : 1,457,861 & 2021 \\
            DeepfakeBench \cite{yan2023deepfakebench} & Deepfake Face Detection & GAN \& DM & $\checkmark$ & 78,195 : 133,102 & 2023 \\
            DeepArt \cite{wang2023benchmarking} & Deepfake Artwork Detection & DM & $\checkmark$ & 64,479 : 73,411 & 2023 \\
            ARIA \cite{li2024adversarial} & Deepfake Artwork Detection & DM & $\times$ & 140,000 : 127,000 & 2024 \\
            CNNSpot \cite{wang2019cnngenerated} & General Image Detection & GAN & $\checkmark$ & 362,000 : 362,000 & 2020 \\
            IEEE VIP Cup \cite{verdoliva2022} & General Image Detection & GAN \& DM & $\times$ & 7,000 : 7,000 & 2022 \\
            DE-FAKE \cite{sha2022fake} & General Image Detection & DM & $\times$ & 20,000 : 60,000 & 2022 \\
            CIFAKE \cite{bird2023cifake} & General Image Detection & DM & $\checkmark$ & 60,000 : 60,000 & 2023 \\
            GenImage \cite{zhu2023genimage} & General Image Detection & GAN \& DM & $\checkmark$ & 1,331,167 : 1,350,000 & 2023 \\
            WildFake \cite{hong2024wildfake} & General Image Detection & GAN \& DM & $\checkmark$ & 2,557,278 : 1,013,446 & 2024
            \\
            \bottomrule[2pt]
        \end{tabular}
    }
    \label{tab:summ_datasets}
\end{table*}

\subsection{Evaluation Metrics} \label{sec:metric}
Faithful evaluation metrics play a crucial role in advancing research. Nevertheless, evaluating AI-generated images is challenging due to the involvement of various attributes that contribute to high-quality generation results, as well as the subjective nature of image evaluation. To ensure reliable assessment, comprehensive metrics should be designed to benchmark synthetic images from two perspectives: 
\begin{enumerate*}[label=(\arabic*)]
    \item the overall quality of generative images;
    \item how incorporating generative images improves downstream tasks.
\end{enumerate*}

Several factors, including text-image alignment, perceptual quality, generation diversity, human preference, and task-specific evaluation, should be taken into account when benchmarking the overall quality of generative images.
For synthetic images from text-guided generation, R-Precision \cite{xu2018attngan}, Captioning Metrics \cite{8578931}, and Semantic Object Accuracy (SOA) \cite{9184960} are widely adopted for assessing the alignment between synthesized images and textual conditions.
Furthermore, it is common to employ some general metrics like Inception Score (IS) \cite{salimans2016improved}, Fréchet Inception Distance (FID) \cite{heusel2017gans}, PSNR, and LPIPS \cite{8578166} for quality and diversity assessment.
Some evaluation metrics are designed to reflect the human preference on T2I generation by fine-tuning the pre-trained CLIP with human preference datasets \cite{wu2023human, kirstain2023pick}, while HYPE \cite{zhou2019hype} is specifically designed for evaluating human face generation.

Evaluation metrics for measuring downstream performance are task-specific. For classification tasks, test accuracy (e.g., top-1 accuracy, top-5 accuracy, Classification Accuracy Scores (CAS) \cite{ravuri2019classification}) has been widely adopted. Recently, Li \emph{et al.} propose the Class-centered Recognition (CLER) \cite{li2023benchmarking} score which is an efficient training-free metric that directly correlates with linear probing CLIP \cite{Radford2021LearningTV} for solving the insufficient correlation of existing metrics (e.g., FID, CLIP score) on downstream classification performance evaluation.
Mean Intersection over Union (mIoU) and Average Precision (AP) are well-suited for semantic segmentation and instance segmentation, respectively.
For object detection tasks, mean Average Precision (mAP) should be the optimal choice for performance evaluation.
It is worth mentioning that, for a thorough and faithful analysis of model performance, task-specific evaluation metrics should be used in combination with other metrics. We tabulate the detailed summary in \Cref{tab:summ_metrics}.

\begin{table}[t]
    \centering
    \caption{Summarization of evaluation metrics used in AIGS. The upper part denotes the evaluation metrics for image synthesis. The lower part denotes the evaluation metrics for measuring the improvement of generative images over downstream tasks.}
    \resizebox{0.9\linewidth}{!}{
        \begin{tabular}{l|c}
            \toprule[2pt] \textbf{Metric} & \textbf{Evaluation Type} \\
            \midrule\midrule
            R-Precision \cite{xu2018attngan} & Text-image Alignment \\
            Captioning Metrics \cite{8578931} & Text-image Alignment \\
            SOA \cite{9184960} & Text-image Alignment \\
            CLIP Score \cite{Radford2021LearningTV} & Text-image Alignment \\
            IS \cite{salimans2016improved} & Perceptual Quality \\
            FID \cite{heusel2017gans} & Perceptual Quality \\
            PSNR & Perceptual Quality \\
            LPIPS \cite{8578166} & Generation Diversity \\
            HPS v2 \cite{wu2023human} & Human Preference \\
            PickScore \cite{kirstain2023pick} & Human Preference \\
            HYPE \cite{zhou2019hype} & Human Face Generation \\
            \midrule
            Test Accuracy \cite{li2023benchmarking} & Classification Improvement \\
            CLER \cite{li2023benchmarking} & Classification Improvement \\
            mIoU \cite{li2023benchmarking} & Semantic Segmentation Improvement \\
            AP \cite{li2023benchmarking} & Instance Segmentation Improvement \\
            mAP \cite{li2023benchmarking} & Detection Improvement \\
            \bottomrule[2pt]
        \end{tabular}
    }
    \label{tab:summ_metrics}
\end{table}

\subsection{Experimental Results}
In this section, we quantitatively benchmark model performance when leveraging synthetic images as training data to perform three types of recognition tasks (i.e., classification, segmentation, and detection), with the evaluation metrics discussed in \Cref{sec:metric}.

It can be seen from \Cref{tab:cls_res} that, for ImageNet classification, models trained solely on synthetic images perform worse than models trained on real images. Nevertheless, augmenting the real images with images generated from the fine-tuned diffusion model imparts a substantial boost in performance across many different classification backbones.
\Cref{tab:seg_res} shows the effectiveness of augmentation using generative images when performing segmentation-based tasks. Leveraging synthetic images as training data alone surprisingly outperforms their real-image counterparts. A significant performance boost can be observed when jointly utilizing both sources of data.
\Cref{tab:det_res} shows how synthetic images benefit the detection model. It is evident that the greatest enhancement has been brought by cut \& paste and foreground object synthesis. The experimental results prove the effectiveness of copy-paste synthesis in a tangible manner.


\begin{table*}[t]
    \centering
    \caption{Comparison of ImageNet \cite{deng2009imagenet} classifier Top-1 Accuracy (\%) performance when generative images are used for data augmentation. The magnitudes of performance boost are highlighted. The results are retrieved from \cite{azizi2023synthetic}.}
    \resizebox{0.9\linewidth}{!}{
        \begin{tabular}{c}
             $\begin{array}{l|c|c|c|c|c|c}
                \toprule[2pt] \text { \textbf{Backbone} } & \text { \textbf{Input Size} } & \text { \textbf{Params (M)} } & \text { \textbf{Real Only} } & \text { \textbf{Generated Only} } & \text { \textbf{Real + Generated} } & \text { \textbf{Performance} } \Delta \\
                \midrule\midrule \multicolumn{7}{c}{\text { ConvNets }} \\
                \hline \text { ResNet-50 } & 224 \times 224 & 36 & 76.39 & 69.24 & 78.17 & \cellcolor{yellow}{+1.78} \\
                \text { ResNet-101 } & 224 \times 224 & 45 & 78.15 & 71.31 & 79.74 & \cellcolor{yellow}{+1.59} \\
                \text { ResNet-152 } & 224 \times 224 & 64 & 78.59 & 72.38 & 80.15 & \cellcolor{yellow}{+1.56} \\
                \text { ResNet-RS-50 } & 160 \times 160 & 36 & 79.10 & 70.72 & 79.97 & \cellcolor{yellow}{+0.87} \\
                \text { ResNet-RS-101 } & 160 \times 160 & 64 & 80.11 & 72.73 & 80.89 & \cellcolor{yellow}{+0.78} \\
                \text { ResNet-RS-101 } & 190 \times 190 & 64 & 81.29 & 73.63 & 81.80 & \cellcolor{yellow}{+0.51} \\
                \text { ResNet-RS-152 } & 224 \times 224 & 87 & 82.81 & 74.46 & 83.10 & \cellcolor{yellow}{+0.29} \\
                \hline \multicolumn{7}{c}{\text { Transformers }} \\
                \hline \text { ViT-S/16 } & 224 \times 224 & 22 & 79.89 & 71.88 & 81.00 & \cellcolor{yellow}{+1.11} \\
                \text { DeiT-S } & 224 \times 224 & 22 & 78.97 & 72.26 & 80.49 & \cellcolor{yellow}{+1.52} \\
                \text { DeiT-B } & 224 \times 224 & 87 & 81.79 & 74.55 & 82.84 & \cellcolor{yellow}{+1.04} \\
                \text { DeiT-B } & 384 \times 384 & 87 & 83.16 & 75.45 & 83.75 & \cellcolor{yellow}{+0.59} \\
                \text { DeiT-L } & 224 \times 224 & 307 & 82.22 & 74.60 & 83.05 & \cellcolor{yellow}{+0.83} \\
                \bottomrule[2pt]
            \end{array}$
        \end{tabular}
    }
    \label{tab:cls_res}
\end{table*}

\begin{table}[t]
    \centering
    \caption{Comparison of semantic segmentation (mIoU) and instance segmentation (AP) performance when generative images are used for data augmentation. The magnitudes of performance boost are highlighted. The results are retrieved from \cite{wu2023datasetdm}.}
    \resizebox{\linewidth}{!}{
        \begin{tabular}{c}
             $\begin{array}{l|c|c|c|c}
                \toprule[2pt] \text { \textbf{Backbone} } & \text { \textbf{Real Only} } & \text { \textbf{Generated Only} } & \text { \textbf{Real + Generated} } & \text { \textbf{Performance} } \Delta \\
                \midrule\midrule \multicolumn{5}{c}{\text { Semantic Segmentation on PASCAL VOC \cite{everingham2010pascal} }} \\
                \hline \text { ResNet-50 } & 43.4 & 60.3 & 66.1 & \cellcolor{yellow}{+22.7} \\
                \text { Swin-B } & 65.2 & 73.7 & 78.5 & \cellcolor{yellow}{+13.3} \\
                \hline \multicolumn{5}{c}{\text { Instance Segmentation on COCO \cite{lin2014microsoft} }} \\
                \hline \text { ResNet-50 } & 4.4 & 12.2 & 14.8 & \cellcolor{yellow}{+10.4} \\
                \text { Swin-B } & 11.3 & 17.6 & 23.3 & \cellcolor{yellow}{+12} \\
                \bottomrule[2pt]
            \end{array}$
        \end{tabular}
    }
    \label{tab:seg_res}
\end{table}

\begin{table*}[t]
    \centering
    \caption{Comparison of COCO detector (mAP@50, mAP) performance when generative images are used for data augmentation. The magnitudes of performance boost are highlighted. The results are retrieved from \cite{ge2022dall}.}
    \resizebox{0.95\linewidth}{!}{
        \begin{tabular}{c}
             $\begin{array}{l|c|c|c|c|c|c}
                \toprule[2pt] \text { \textbf{Metric} } & \text { \textbf{Real Only} } & \text { \textbf{Real Only w/ Cut Paste} } & \text { \textbf{Generated Foreground} } & \text { \textbf{Generated Only} } & \text { \textbf{Real + Generated} } & \text { \textbf{Performance} } \Delta \\
                \midrule\midrule \multicolumn{7}{c}{\text { Object Detection on PASCAL VOC \cite{everingham2010pascal} }} \\
                \hline \text { mAP@50 } & 9.12 & 29.60 & 48.14 & 44.59 & 51.82 & \cellcolor{yellow}{+42.7} \\
                \text { mAP } & 2.35 & 10.82 & 21.62 & 20.72 & 25.87 & \cellcolor{yellow}{+23.52} \\
                \hline \multicolumn{7}{c}{\text { Object Detection on COCO \cite{lin2014microsoft} }} \\
                \hline \text { mAP@50 } & 1.47 & 2.89 & 17.87 & 16.22 & 20.82 & \cellcolor{yellow}{+19.35} \\
                \text { mAP } & 0.92 & 1.23 & 8.64 & 7.66 & 10.63 & \cellcolor{yellow}{+9.71} \\
                \bottomrule[2pt]
            \end{array}$
        \end{tabular}
    }
    \label{tab:det_res}
\end{table*}



\section{Social Impacts} \label{sec:social}
In recent years, the hot concept of \textit{AI}-\textit{G}enerated \textit{C}ontent (AIGC) has gained immense attention and interest. The surge in AIGS offers pioneering insights in formulating novel training frameworks and enhancing the performance of computer vision models, which has influenced and will keep renovating our society in both positive and potentially negative aspects. In this section, we will discuss the correlation between AIGS and AIGC (\Cref{sec:aigc}), and dissect the underlying social impacts into trustworthiness (\Cref{sec:trust}), misuse (\Cref{sec:misuse}), as well as the environmental influence (\Cref{sec:envir}) of AIGS.

\subsection{Correlation with AIGC} \label{sec:aigc} 

\begin{figure}[ht]
    \centering
    \includegraphics[width=0.75\linewidth]{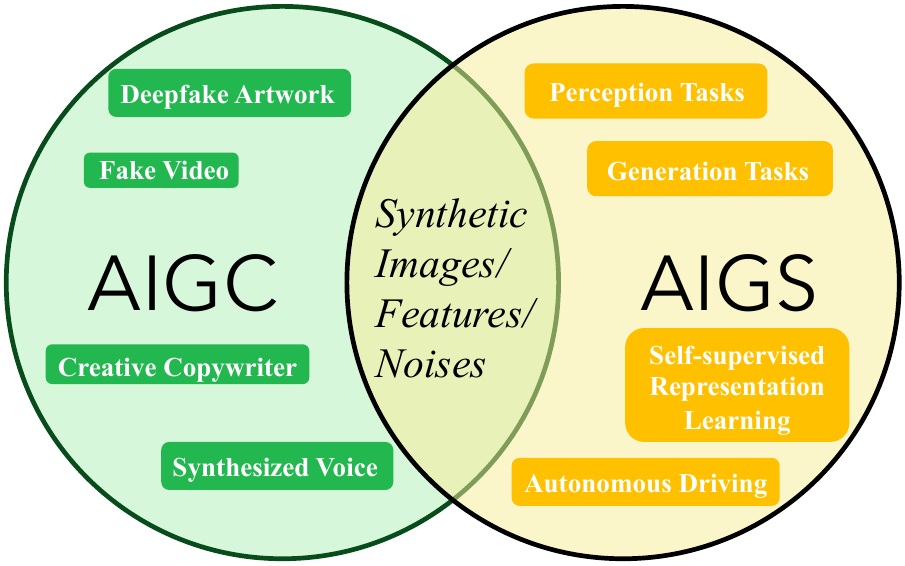}
    \caption{Correlation of AIGS with AIGC.}
    \label{fig:corr}
\end{figure}

Accompanied by the emergence of ChatGPT and Stable Diffusion, the research topics related to AIGC have become unprecedentedly popular. AIGS has strong correlation with AIGC, since they both use deep learning techniques to generate novel content.
More specifically, synthetic products (e.g., synthetic images, synthetic features \cite{xu2023open}, and synthetic noises \cite{raj2023_dreambooth3d}) are what they have in common.

Nevertheless, AIGS is designed to use these synthetic products as data sources for downstream applications with an emphasis on computer vision tasks, whereas AIGC aims to produce a broader range of creative works including but not limited to visual content, text content, audio content, etc. 
Detailed visualization of their correlation can be viewed from \Cref{fig:corr}.

\subsection{Trustworthiness} \label{sec:trust}
The trustworthiness of AI has drawn a tremendous amount of attention in recent years, emphasizing the need of bypassing the side effects that AI could bring in. As a large and complex subject, trustworthy AI has multiple dimensions \cite{liu2022trustworthy}, including robustness, fairness, explainability, privacy protection, accountability, etc. Specifically, we will focus on the following three aspects to expound the positive social influences of AIGS.

\textbf{Privacy Protection.} \label{sec:privacy}
One of the biggest motivations for replacing real data with synthetic data during training is for reducing privacy concerns, especially when dealing with sensitive human data, such as medical data, human faces, etc. To alleviate the privacy issues, people can use synthetic data from AI models to perform the same tasks without violating the privacy protection act. Many studies \cite{8411200, Zhang2021OnTA, Bozkir2019PrivacyPG, hillerstrom2014generation} have illustrated that synthetic data with a similar distribution and characteristics to the real data can be efficiently generated and effectively utilized.

\textbf{Fairness.} \label{sec:fairness}
Synthetic data that reflects the underlying statistical properties tends to inherit the bias from the real data \cite{lu2023machine}. Theoretically, three approaches have been explored to address this issue:
\begin{enumerate*}[label=(\arabic*)]
    \item data pre-processing such as massaging, reweighting, and sampling;
    \item data in-processing that introduces fairness evaluation during the learning progress;
    \item data post-processing that modifies the predicted results after inference time.
\end{enumerate*}
Several existing studies \cite{9025595, Zhai2021DemodalizingFR, 8575453, 9507539, mcduff2021synthetic} demonstrate the capability of synthetic data in training an unbiased model.

\textbf{Robustness.} \label{sec:robust}
In the context of the trustworthy AI, robust synthetic data refers to the data that accurately represents the underlying distribution of the real data and retains its statistical properties. Diverse and representative synthetic data samples span the entire data distribution, making the model better equipped to learn from a wide range of scenarios. In addition, the model can better adapt to new and unseen data with real-world distributions, thanks to the enhanced generalization ability endowed by generative data augmentation.

\subsection{Misuse} \label{sec:misuse}
One notable downside of AIGS lies in generating or manipulating images for malicious and illegal purposes. To be more specific, one may pose negative social impacts by spreading pornography, violence, and fraud information via synthetic images. To prevent such misuses, people can adopt detection techniques (e.g., metadata analysis, watermark detection, contextual analysis, etc.) to identify the adverse AI-generated images. In the meantime, a robust safeguard mechanism, labeling, as well as access control should be carefully taken into account when deploying AIGS applications, to minimize the chance of misusing synthetic data.

\subsection{Environment} \label{sec:envir}
As neural image synthesis involves numerous deep-learning-based algorithms, current generative models and neural rendering models require a mass of computational resources to train, and therefore cause huge energy consumption. This is harmful to the environment and results in the global warming before the large-scale employment of renewable energy.
One possible solution to reduce the demand for GPU resources is relevant to model generalization. For instance, one can leverage prior knowledge in large and pre-trained foundation models to drastically accelerate the training processes of various downstream tasks.

\section{Challenges \& Discussion} \label{sec:challenge}
While we have witnessed significant progress in various AIGS paradigms, several open challenges remain for future exploration. In this section, we overview four major challenges regarding the explainability (\Cref{sec:xai}), evaluation metrics (\Cref{sec:metric_challen}), model selection (\Cref{sec:choice}), and 3D awareness (\Cref{sec:3daware}) when performing AIGS.

\subsection{Towards Faithful Explainability} \label{sec:xai}
In \Cref{sec:trust}, we discussed three positive trustworthy AI attributes that AIGS holds. However, the explainability of AIGS remains weak. Even though the AIGS-based approaches yield smaller domain gaps \cite{joshi2022synthetic} compared to their simulation-based (i.e., rendering images from 3D graphics engines) counterparts, current AIGS methods cannot explain corner cases and extreme outliers presented in the original data distribution while the applied domain spreads wider.

According to Huang \emph{et al.} \cite{xai}, exploring outliers and their influences on the parameterization of existing approaches could be a promising research direction. Future efforts are needed for building up trustworthy AIGS systems with more comprehensive explainability.

\subsection{Towards Precise Evaluation Metrics} \label{sec:metric_challen}
The research on the evaluation of AI-generated images remains open. Common quantitative metrics are sometimes task-specific or even biased. For example, FID is bounded by the pre-trained datasets, which poses domain gaps with various target datasets.
Qualitative evaluation measures are not absolutely reliable either. For instance, in a user study, human thoughts are adopted for synthesis quality assessment, whereas this may easily lead to subjective feedback. Designing precise and faithful evaluation metrics is indispensable and crucial for future AIGS development.

Considering the widespread applications of T2I generative models, tools for cross-modal alignment measurement, like the pre-trained CLIP \cite{Radford2021LearningTV}, can be leveraged to evaluate the proximity between the generated images and the text prompts.
Moreover, the newly proposed DreamSim \cite{fu2023learning} exemplifies a novel perceptual metric for human visual similarity. It computes feature-level cosine distances based on embedding backbones pre-trained on synthetic data.
Ongoing exploration of such evaluation metrics is beneficial for the evolution of AIGS.

\subsection{Towards Elaborated Model Choice} \label{sec:choice}
The inductive biases from underlying generative models might not 
be apparent \cite{bhanot2021problem}. Misleading factors such as sample selection biases and class imbalances can contribute to these drawbacks.
Furthermore, though DMs achieve higher-resolution and photorealistic synthesis, their inference speeds are significantly slower compared to GAN-based AIGS methods. To this end, some work \cite{dockhorn2022score} has explored how to speed up the DMs.

On the other hand, CNN-based GANs lack capabilities of dealing with multi-modal inputs. To this end, several studies \cite{9665852, zhang2022styleswin} manage to mitigate these issues through applying Transformer-based \cite{vaswani2017attention} architecture in GANs. Nonetheless, careful and thoughtful generator selection remains essential along this research line.

\subsection{Towards 3D Awareness} \label{sec:3daware}
With the rise of neural rendering models, especially NeRF and 3DGS, the adaptation to 3D-aware AIGS could be the next transformative breakthrough since it allows the model to incorporate and interpret 3D geometry of the real world.
However, achieving reliable 3D understanding remains challenging, as current approaches still struggle to reconstruct consistent geometry and appearance, particularly under unposed or sparse-view conditions where geometric cues are limited.


Moreover, when using synthesized views as augmented data, the balance between real and generated samples requires careful consideration. A promising direction is to incorporate stronger prior knowledge \cite{3dgp} and introduce scene-specific supervision \cite{yen2022nerfsupervision} to mitigate these limitations. Once 3D-aware AIGS frameworks can effectively handle complex natural scenes, they will unlock a broader range of real-world applications.

\section{Conclusion} \label{sec:conslusion}
This survey has covered main technologies and applications for AI-generated images as data sources.
Especially, we introduce models for neural image synthesis, including Generative Adversarial Networks, diffusion models, and neural radiance fields.
After that, we discuss AIGS methodologies for automatic label acquisition and dataset augmentation.
Moreover, we explore the enormous potential of AIGS paradigms on energizing various applications such as visual perception and visual generation tasks, self-supervised learning, robotics, as well as autonomous driving.
We also conduct an extensive investigation of existing synthetic datasets and evaluation metrics for AIGS, with a tabularized summary and experimental results.
Last but not least, we present our humble insights on the social impacts and open challenges of current AIGS, supported by real-world examples.
While challenges persist, we believe that the potential of AIGS has not been fully activated.
Future research and development of AIGS methodologies can further reinforce the functionality and reliability of AI-generated data.

{\small

\input{main.bbl}
}

\patchcmd{\IEEEbiographynophoto}
  {\interlinepenalty500%
\vskip 4\baselineskip plus 1fil minus 0\baselineskip}
  {\interlinepenalty10000%
\vskip 2\baselineskip}
  {}
  {\PackageError{biography-spacing}{Unable to set fixed biography spacing}{Check the IEEEtran biography definition.}}
\raggedbottom
\begin{IEEEbiographynophoto}{Zuhao Yang} is currently a Ph.D. candidate with the Visual Intelligence Lab at Nanyang Technological University, Singapore, under the supervision of Shijian Lu. He received the B.S. degree in computing science from the University of Alberta. He previously worked on native multimodal foundation models at Kimi (Moonshot AI) and collaborated with researchers at MiroMind and ByteDance. His research focuses on video-centric multimodal intelligence, including temporal grounding, agentic reasoning, and long-horizon tool use.
\end{IEEEbiographynophoto}

\begin{IEEEbiographynophoto}{Yu Wei} is a Ph.D. student at the College of Computing and Data Science, Nanyang Technological University, Singapore. He is with the Visual Intelligence Lab. His research interests include 3D vision and deep learning, particularly 3D scene reconstruction and understanding.
\end{IEEEbiographynophoto}

\begin{IEEEbiographynophoto}{Fangneng Zhan} is an Assistant Professor at The Hong Kong University of Science and Technology, where he leads the World Mind Lab. He is also Chief Scientist of Simate AI. He received the Ph.D. in computer science and engineering from Nanyang Technological University in 2021. He previously held postdoctoral positions at Harvard University and the Max Planck Institute for Informatics and was a researcher at MIT. His research covers world models, robotics, and physical agents, including 3D and multimodal modeling, robot learning, and autonomous perception.
\end{IEEEbiographynophoto}

\begin{IEEEbiographynophoto}{Kunhao Liu} received the bachelor's degree in software engineering from Beihang University, Beijing, China, in 2022. He received the Ph.D. degree in computer science from Nanyang Technological University, Singapore, in 2026, under the supervision of Shijian Lu. His research focuses on generative vision, particularly video foundation models and world models. He was a visiting student at the Max Planck Institute for Informatics and a research intern at Tencent ARC Lab and morpheus.ai.
\end{IEEEbiographynophoto}

\begin{IEEEbiographynophoto}{Muyu Xu} is currently a Ph.D. candidate at Nanyang Technological University, Singapore. His research interests include 3D vision and deep learning. His work spans neural radiance fields, 3D Gaussian splatting, and novel view synthesis.
\end{IEEEbiographynophoto}

\begin{IEEEbiographynophoto}{Xiaoqin Zhang} received the bachelor's degree from Central South University, China, in 2005, and the Ph.D. degree from the Institute of Automation, Chinese Academy of Sciences, Beijing, China, in 2010. He is currently a Full Professor at Zhejiang University of Technology, China. He previously held faculty positions at Wenzhou University. His research interests include artificial intelligence, computer vision, machine learning, and image processing. He was a recipient of China's State Natural Science Award (Second Class) for 2020 and is an IET Fellow.
\end{IEEEbiographynophoto}

\begin{IEEEbiographynophoto}{Ling Shao} (Fellow, IEEE) received the B.Eng. degree from the University of Science and Technology of China and the M.Sc. and D.Phil. degrees from the University of Oxford, U.K. He is the Global President and Chief AI Officer of Terminus Group. He was the founding CEO of the Inception Institute of Artificial Intelligence and the founding Provost and Executive Vice President of Mohamed bin Zayed University of Artificial Intelligence. His research interests include computer vision, generative AI, multimodal AI, and AI for healthcare. He is a Fellow of the IAPR and received the Mohammed bin Rashid Medal for Scientific Distinction.
\end{IEEEbiographynophoto}

\begin{IEEEbiographynophoto}{Shijian Lu} is an Associate Professor with the College of Computing and Data Science, Nanyang Technological University, Singapore. He received the Ph.D. degree in electrical and computer engineering from the National University of Singapore. His research interests include computer vision, deep learning, visual intelligence, and image and video analytics. Before joining NTU, he held research and leadership positions at the Institute for Infocomm Research, A*STAR, Singapore. He has held associate editorships for Pattern Recognition and Neurocomputing and area chair or senior program committee roles at computer vision and AI conferences.
\end{IEEEbiographynophoto}

\end{document}

%% file: commands/newcommand.tex
\newcommand{\etal}{\textit{et al.}}